\documentclass[10pt,twocolumn,letterpaper]{article}

\usepackage[pagenumbers]{cvpr}

\usepackage{cuted} %

\usepackage{currfile} %

\usepackage{caption} %
\usepackage[table]{xcolor}
\usepackage{xspace}
\usepackage{tikz}
\usepackage{tabularx}
\usepackage{tocloft}
\usepackage{multirow}
\usepackage{algorithm}
\usepackage[noend]{algpseudocode}

\newcolumntype{Y}{>{\centering\arraybackslash}X}
\newcommand{\Cone}{$s_1@t_1$\xspace}
\newcommand{\Ctwo}{$s_2@t_2$\xspace}
\newcommand{\Ptwo}{$s_1@t_2$\xspace}

\definecolor{bbrow}{gray}{0.93}
\definecolor{oursrow}{RGB}{232,240,254}
\definecolor{refc}{gray}{0.5}

\providecommand{\rep}[1]{#1}
\providecommand{\famsep}[1]{%
  \noalign{\vskip 1pt}%
  \multispan{#1}%
  \cleaders\hbox{\rule{1pt}{\arrayrulewidth}\kern1.5pt}\hfill\kern0pt\cr
  \noalign{\vskip 1.5pt}%
}

\providecommand{\plus}{\quad + }
\providecommand{\hd}[1]{{\bfseries\boldmath #1}}

\providecommand{\rf}[1]{\textcolor{refc}{#1}}
\providecommand{\cm}{\checkmark}

\DeclareRobustCommand{\narrowtexttt}[1]{\scalebox{0.85}[1]{\texttt{#1}}} %
\newcommand{\data}{SCE\xspace}
\newcommand{\ours}{\narrowtexttt{\textbf{StreamLoG}}\xspace}

\makeatletter
\def\@itabbr#1{\def\@itabbr@text{#1}\futurelet\@let@token\@itabbr@}
\def\@itabbr@{\ifx\@let@token.\textit{\@itabbr@text}\else\textit{\@itabbr@text.}\null\fi\xspace}
\DeclareRobustCommand\eg{\@itabbr{e.g}}
\DeclareRobustCommand\Eg{\@itabbr{E.g}}
\DeclareRobustCommand\ie{\@itabbr{i.e}}
\DeclareRobustCommand\Ie{\@itabbr{I.e}}
\makeatother

\definecolor{cvprblue}{rgb}{0.21,0.49,0.74}
\usepackage[pagebackref,breaklinks,colorlinks,allcolors=cvprblue]{hyperref}

\def\confName{CVPR}
\def\confYear{2027}
\title{Revisiting Compact Streaming Video Memory: Preserving Both Past and Present}

\author{
    Hongyeob Kim\textsuperscript{1}\hspace{0.25em}
    Seunggyu Kim\textsuperscript{1}\hspace{0.25em}
    Suyeon Hwang\textsuperscript{2}\hspace{0.25em}
    Tae-Hyun Oh\textsuperscript{3} \hspace{0.25em}
    Sungeun Hong\textsuperscript{1,2}\thanks{Corresponding Author.} \\
    \textsuperscript{1}Dept. of Applied Artificial Intelligence of Sungkyunkwan University \\
    \textsuperscript{2}Dept. of Immersive Media Engineering of Sungkyunkwan University \quad
    \textsuperscript{3}KAIST \\ [0.3em]
    \hypersetup{urlcolor=magenta}\url{https://aim-skku.github.io/StreamLoG}
    \vspace{-3mm}
}

\begin{document}
\maketitle
\begin{abstract}
In streaming video understanding, questions may ask about either the current scene or an earlier moment; \eg, after a drive from day to night, the answer to ``Is it daytime?'' is \emph{`no'} for the current scene but \emph{`yes'} for the beginning of the drive.
Yet existing benchmarks evaluate state tracking and history recall separately.
To address this, we introduce State-Contrastive Evaluation (SCE), a complementary protocol that jointly scores prior- and current-state questions, testing whether a memory keeps both.
With SCE, we find that bounded-memory methods, which discard or merge the past, often lose prior states, whereas retrieval-based methods, which read the past for every question, let it interfere with current-state answers.
We therefore propose \ours, a training-free memory that decouples what to keep from when to read.
It stores each past frame as compact low-rank residuals over shared statistics, keeping every frame retrievable.
It looks back only when the recent frames barely change the model's representation of the question, limiting interference with current-state answers.
On SCE, \ours answers both prior- and current-state questions at the same moment correctly in 22.0\% of pairs, versus 15.2\% for the strongest method, while a recent-frame baseline strong on existing benchmarks does so in only 4.8\%, below a text-only model.
On three existing benchmarks, ours remains competitive while using about $1/16$ of the uncompressed KV memory per frame in FP16.
\end{abstract}
\vspace{-3mm}
    
\section{Introduction}
\label{sec:intro}

\begin{figure}[t]
    \centering
    \begin{subfigure}{\linewidth}\centering
    \includegraphics[width=\linewidth]{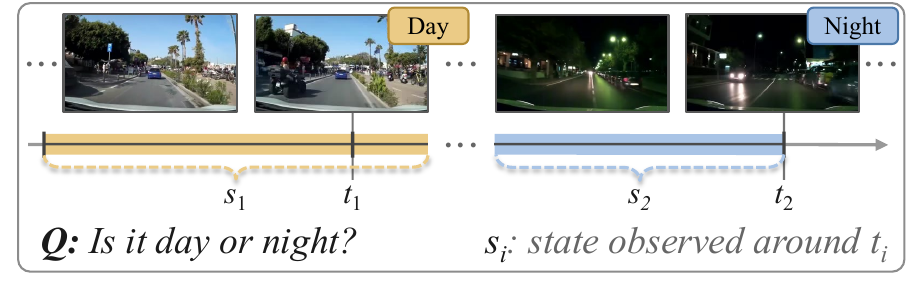}\par
    \vspace{-2mm}
    {\small (a) Input stream and question}
    \phantomsubcaption\label{fig:teaser_a}
    \end{subfigure}\par\vspace{2pt}
    \begin{subfigure}{\linewidth}\centering
    \includegraphics[width=\linewidth]{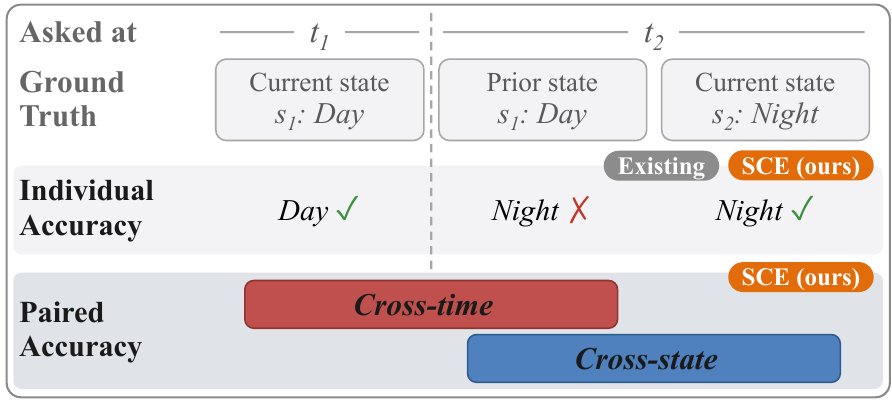}\par
    {\small (b) Individual vs.\ paired scoring}
    \phantomsubcaption\label{fig:teaser_b}
    \end{subfigure}
    \caption{\textbf{State-contrastive evaluation of streaming memories.} (a) The scene changes from $s_1$ at $t_1$ to $s_2$ by $t_2$; $s@t$ denotes a question about $s$ asked at $t$. (b) Most existing benchmarks score each answer alone, whereas \data counts a pair only if both answers are correct. \emph{Cross-time} pairs \Cone with \Ptwo, to assess recall after the change; \emph{Cross-state} pairs \Ptwo with \Ctwo, to assess whether the two states are told apart. The example gets two of three answers right but fails both pairs.}
    \vspace{-3mm}
    \label{fig:teaser}
\end{figure}

Streaming video understanding aims to answer questions about an ongoing video at any moment from what has been observed so far~\cite{videollmonline2024,flashvstream2025}.
Recent works~\cite{dispider2025,streamingvlm2026} extend video large language models (Video-LLMs)~\cite{qwen25vl2025,llavaonevision2025,llavavideo2025} to this setting, caching the keys and values (KV) of past frames as memory.
Questions arrive without warning~\cite{streammem2025,protokv2026} and may concern the present or any earlier moment.
As scenes change, the same question can have different answers at different times; \eg, after driving from day to night, the answer to ``Is it day or night?'' is `\emph{night}' for the current state at $t_2$ but `\emph{day}' for the prior state at $t_1$ (see \Cref{fig:teaser}).
Therefore, streaming models need to track the current state while still recalling the history.

State changes thus pose the central challenge for streaming memory: past and present evidence points to different answers, and a model has to keep both and tell them apart.
Existing streaming benchmarks~\cite{ovobench2025,streamingbench2026,rtvbench2025}, however, evaluate questions about the present and the past separately.
A correct answer therefore does not reveal whether a model can both track the updated state and recall the earlier one.
Recent analyses show that models given only the most recent frames rival memory-based methods on OVO-Bench and StreamingBench~\cite{simplestream2026,deltas2026}.
RTV-Bench~\cite{rtvbench2025} tracks answers that change over time and scores related questions jointly, but only conditions an advanced question on basic ones.
This leaves open whether a model can follow a state change without losing the state before it.

To fill this gap, we introduce State-Contrastive Evaluation (\data), a benchmark built from OVO-Bench~\cite{ovobench2025}, StreamingBench~\cite{streamingbench2026}, and RTV-Bench~\cite{rtvbench2025}.
Its key idea is to pair questions (\Cref{fig:teaser_b}) about the past and the present either at different query times (\emph{Cross-time}: $s_1@t_1$ and $s_1@t_2$) or at the same query time (\emph{Cross-state}: $s_1@t_2$ and $s_2@t_2$).
Through these paired questions, \data separates three abilities that per-question accuracy conflates: perceiving a state, recalling it after it changes, and distinguishing it from the current state.

\begin{figure}[t]
    \centering
    \includegraphics[width=0.98\linewidth]{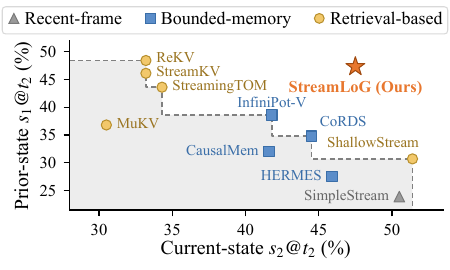}
    \vspace{-2mm}
    \caption{\textbf{Current- vs.\ prior-state accuracy on \data with LLaVA-OneVision.} Existing methods favor either the present or the past, whereas \ours lies beyond their frontier (dashed).}
    \label{fig:tradeoff}
    \vspace{-3mm}
\end{figure}

\data exposes what per-question scores in existing benchmarks have hidden.
A recent-frame baseline, SimpleStream~\cite{simplestream2026}, competitive on existing benchmarks, answers both \emph{Cross-state} questions correctly in only 4.8\% of pairs, below a text-only model.
Memory-based methods, which either bound or archive the growing KV cache, do better but fall short in opposite directions (\Cref{fig:tradeoff}).
Bounded-memory methods~\cite{infinipotv2025,hermes2026,causalmem2026,cords2026} prune or merge tokens to keep a fixed-size cache, losing evidence of earlier states.
Retrieval-based methods~\cite{rekv2025,streamkv2026,mukv2026,streamingtom2026} archive past KV and recall prior states better, but reading the archive per question lets the past intrude on current-state answers.

Both shortcomings stem from tying what a memory \emph{keeps} to how it is \emph{used}: bounded memories forget earlier states to stay small, and retrieval memories read their archive for every question.
We argue for decoupling the two: \emph{keeping every frame at a small cost and looking back only when the present cannot answer the question.}
Based on this, we propose \ours (\textbf{Stream}ing \textbf{Lo}w-rank residuals with \textbf{G}ated recall), a training-free streaming memory.

Since consecutive frames largely overlap, our method only stores what each frame adds with low-rank representation.
Unlike prior retrieval-based methods, which keep fewer tokens or lower their precision~\cite{mukv2026,streamkv2026,streamingtom2026}, our method further compresses the feature dimension and shares recurring statistics across frames, keeping each frame retrievable. 
To decide when to look back, the look-back gate compares the model’s representation of the question with and without access to recent frames.
A small change suggests that the recent frames lack the answer, so the gate looks back.

On \data, \ours achieves the best \emph{Cross-state} accuracy, reaching 22.0\% versus 15.2\% for the strongest baseline on LLaVA-OneVision.
At matched memory, it outperforms token selection, merging, and quantization-based compression, and even surpasses the uncompressed cache methods on \emph{Cross-state} (\Cref{tab:compression_swap}).
It also remains competitive on OVO-Bench~\cite{ovobench2025}, StreamingBench~\cite{streamingbench2026}, and RTV-Bench~\cite{rtvbench2025} with about $1/16$ of raw KV per frame in FP16.

Our contributions are summarized as follows:
\begin{itemize}
    \item We introduce \data, a benchmark that pairs prior- and current-state questions on the same or different observations to test whether a streaming memory preserves both.
    \item We propose \ours, a training-free memory that decouples keeping the past from reading it, storing only what each frame adds and looking back only when the recent frames lack the answer.
    \item Evaluating ten streaming methods from three families, we uncover a trade-off between perceiving the present and recalling the past, and show that decoupling what to keep from when to read overcomes this trade-off (\Cref{fig:tradeoff}).
\end{itemize}

\section{Related Work}
\label{sec:related}

\subsection{Streaming Video Understanding}
Streaming Video-LLMs extend offline models~\cite{llavaonevision2025,llavavideo2025,qwen25vl2025} to answer questions while a video is being observed~\cite{videollmonline2024,flashvstream2025,dispider2025,streamingvlm2026,timechatonline2025,streamready2026}.
Their answers must track visual changes~\cite{streammind2025,streambridge2025,vispeak2025}.
Streaming benchmarks pose questions throughout a stream, each answered from the frames observed so far~\cite{ovobench2025,streamingbench2026,streamarena2026}.
Related benchmarks further track time-varying answers~\cite{rtvbench2025} or stateful counts~\cite{svcbench2026}.
However, recent frames alone often suffice for strong scores~\cite{simplestream2026,deltas2026}, and adding memory can trade current-scene perception for long-range recall~\cite{simplestream2026}.
Our \data measures this trade-off directly by asking about both states of the same change at one query time and scoring them as a pair.

\subsection{Memory for Streaming Video-LLMs}
Bounded-memory methods keep a fixed budget through KV eviction~\cite{dscache2026}, token selection~\cite{infinipotv2025,cords2026,causalmem2026}, selection with frame-wise KV merging~\cite{streammem2025}, hierarchical memory~\cite{hermes2026,curvestream2026}, or summarization~\cite{protokv2026}.
This bounds memory at the cost of earlier evidence.
Even those using low-rank bases or residual statistics~\cite{causalmem2026,protokv2026} discard tokens or fold the past into a fixed summary.
Retrieval-based methods instead maintain a growing archive of past KV~\cite{rekv2025}.
To curb its growth, they merge frames and select KV blocks~\cite{streamkv2026}, select tokens at multiple granularities~\cite{mukv2026}, or combine token reduction with quantization~\cite{streamingtom2026}.
Yet each retained token keeps its full feature dimension (\Cref{tab:compare}).

For language and vision-language models, KV caches have been compressed with low-rank projections~\cite{eigenattention2024,palu2025}, cross-layer bases~\cite{xkv2026}, Tucker decomposition~\cite{jolt2026}, multi-head factorization~\cite{attentionpack2026}, cluster centroids~\cite{centroidkv2026}, and residual codes~\cite{residualkv2026}.
These methods assume a fixed context that can be calibrated or factorized all at once.
A stream, in contrast, has to be compressed frame by frame, before any question arrives.
More fundamentally, these approaches decide what to discard from the cache.
A stream suggests a different view:
consecutive frames largely repeat one another, so the past can be kept by storing only what each frame adds, without discarding any frame.
\ours builds on this view while keeping every frame individually retrievable.

\subsection{Deciding When to Retrieve}
Retrieval methods select stored frames or KV blocks by their relevance to the question~\cite{rekv2025,streamkv2026,livevlm2026,cacheflow2025}.
Retrieving for every question, however, can interfere with current-state answers.
ShallowStream~\cite{shallowstream2026} and concurrent FlashBack~\cite{flashback2026} decide from the question text alone.
Other methods rely on uncertainty from an external encoder~\cite{wrws2026} or an initial pass~\cite{oasis2026,weavetime2026}, or on whether the question directly matches an entity memory~\cite{folio2026}.
Others learn a sufficiency predictor~\cite{spare2026} or a policy for expanding the visual context~\cite{streamscout2026}.
Yet whether the past is needed depends on what the present already shows, not on the question alone.
The same question may be answered by the current frames at one moment and require the past at another.
Our gate instead measures whether the recent frames change the model's representation of the question.
This requires no training, first-pass answer, or external encoder.

\section{Method}
\label{sec:method}

\noindent\textbf{Problem setup.} A frozen Video-LLM receives a video stream frame by frame and encodes each frame into $N$ visual tokens, whose keys and values are cached at every layer.
Questions $q$ arrive at query times $t$ and are answered using only the frames observed so far.
Since full KV storage grows linearly with stream length, the model must compress past observations without knowing which evidence later questions will require.
To evaluate how well a streaming memory supports both current-state tracking and prior-state recall, we introduce \data{} (\Cref{sec:sce}). 
\Cref{sec:factor,sec:gate,sec:readout} describe how we compress past KV, decide when to look back, and read the retrieved records.

\subsection{State-Contrastive Evaluation}
\label{sec:sce}
We derive \data from OVO-Bench~\cite{ovobench2025}, StreamingBench~\cite{streamingbench2026}, and RTV-Bench~\cite{rtvbench2025}, which include questions repeated at multiple query times within a video.
Using the question text and source annotations, we identify pairs of query times $t_1$ and $t_2$ at least $5\,\mathrm{s}$ apart at which the same question has different answers, $s_1$ and $s_2$, respectively.
For each video, we select and manually review the pair with the largest time gap $t_2-t_1$.
For each selected pair, we start the clip at $t_1-\ell$ and stream it causally up to $t_2$.
The \emph{pre-query context length} $\ell$ defaults to $8\,\mathrm{s}$, providing visual context before the first query while keeping $t_1$ near the start of the clip.
At $t_1$ and $t_2$, we ask about the current state (\Cone or \Ctwo) with the source questions and the original answer choices.
Since each question refers to the state at its query time, we remove
redundant phrases such as ``at this moment'' from these questions.
At $t_2$, we also ask a prior-state question (\Ptwo) with the prefix ``Near the beginning of this clip'' to refer back to $s_1$.
We include $s_1$ as the correct answer and $s_2$ as a distractor in \Ptwo{}, using four answer choices by default.
In \Cref{sec:sensitivity}, we examine how varying $\ell$ and the wording of the \Ptwo{} prefix affects performance.

\begin{table}[t]
\centering
\small
\setlength{\tabcolsep}{4pt}
\renewcommand{\arraystretch}{0.92}
\setlength{\aboverulesep}{0.25ex}\setlength{\belowrulesep}{0.35ex}
\caption{\textbf{Retrieval-based methods.} Token, Chan.\ (channel), Low-bit: fewer stored tokens, dimensions per token, or low precision. Adapt.\ Retrieval: decides per question whether to retrieve.}
\label{tab:compare}
\vspace{-2pt}
\begin{tabular}{lcccc}
\toprule
\multirow{2}{*}[-0.3ex]{\hd{Method}}
  & \multicolumn{3}{c}{\hd{Compression}}
  & \multirow{2}{*}[-0.5ex]{\hd{\shortstack{Adapt.\\[-0.6pt]Retrieval}}} \\
\cmidrule(lr){2-4}
  & \hd{Token} & \hd{Chan.} & \hd{Low-bit} & \\
\midrule
ReKV~\cite{rekv2025} & & & & \\
StreamKV~\cite{streamkv2026} & \cm & & & \\
MuKV~\cite{mukv2026} & \cm & & & \\
StreamingTOM~\cite{streamingtom2026} & \cm & & \cm & \\
ShallowStream~\cite{shallowstream2026} & \cm & & & \cm \\
\rowcolor{oursrow}\ours~\textbf{(Ours)} & \cm & \cm & \cm & \cm \\
\bottomrule
\end{tabular}
\vspace{-3mm}
\end{table}

\newcommand{\figsectag}[1]{{\fontsize{8.5}{9}\selectfont\bfseries\hyperref[#1]{\textcolor{cvprblue}{\underline{Sec.~\ref*{#1}}}}}}
\newcommand{\figsubtag}[1]{{\fontsize{7}{8}\selectfont\bfseries\hyperref[#1]{\textcolor{cvprblue}{\underline{\ref*{#1}}}}}}
\begin{figure*}[t]
  \centering
  \resizebox{0.97\textwidth}{!}{
  \begin{tikzpicture}
    \node[anchor=south west, inner sep=0] (img) at (0,0)
      {\includegraphics[width=\textwidth]{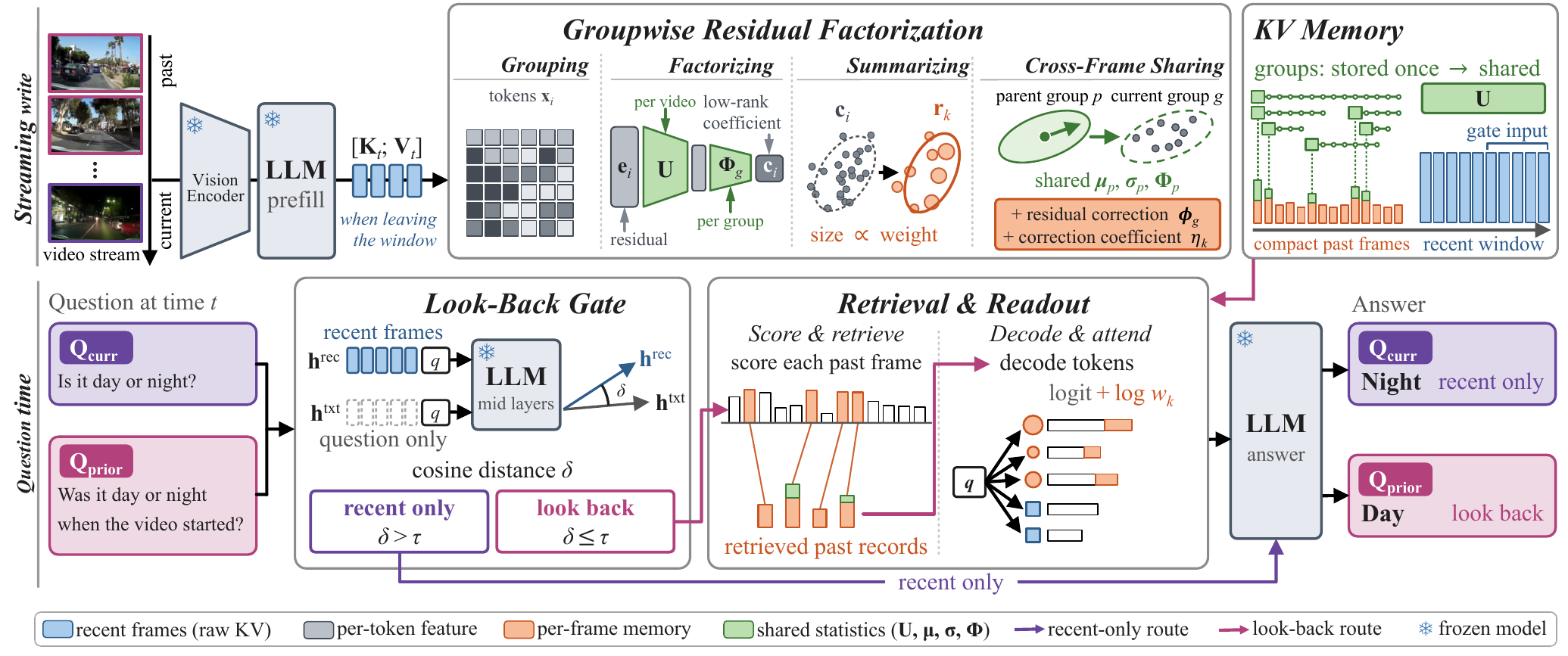}};
    \begin{scope}[x={(img.south east)}, y={(img.north west)}]
      \node[anchor=base east, inner sep=0] at (0.3500, 0.9383) {\figsectag{sec:factor}}; %
      \node[anchor=base east, inner sep=0] at (0.2620, 0.5230) {\figsectag{sec:gate}}; %
      \node[anchor=base east, inner sep=0] at (0.5263, 0.5230) {\figsectag{sec:readout}}; %
      \node[anchor=base east, inner sep=0] at (0.3200, 0.8855) {\figsubtag{sec:group}}; %
      \node[anchor=base east, inner sep=0] at (0.4240, 0.8855) {\figsubtag{sec:factorize}}; %
      \node[anchor=base east, inner sep=0] at (0.5390, 0.8855) {\figsubtag{sec:summarize}}; %
      \node[anchor=base east, inner sep=0] at (0.6530, 0.8855) {\figsubtag{sec:reuse}}; %
    \end{scope}
  \end{tikzpicture}}
\caption{\textbf{Overview of \ours.} (i)~Each frame is prefilled by the frozen Video-LLM, and the latest $W$ frames remain uncompressed. (ii)~When it leaves the window, its tokens are grouped, and each group keeps a center and encodes its residuals ($e_i$) through $\mathbf{U}$ and $\boldsymbol{\Phi}_g$ into a few weighted summary tokens ($c_i \rightarrow r_k$). $\mathbf{U}$ is shared across the video, and $(\boldsymbol{\mu},\boldsymbol{\sigma},\boldsymbol{\Phi})$ across frames whose groups match. (iii) At question time, the gate compares the question representations with and without access to the most recent frames. If \(\delta > \tau\), the model answers from those frames alone. (iv)~Otherwise, it retrieves and decodes the most relevant past records for weighted attention.}
\label{fig:method}
\vspace{-3mm}
\end{figure*}

\subsection{Groupwise Residual Factorization}
\label{sec:factor}
Following ReKV~\cite{rekv2025}, \ours uses its default frame budget  (\ie, 64 frames) for video prefill.
For each incoming frame, our method uses uncompressed KV from the $W$-frame recent window and fills the remaining frame budget with decoded KV from the most recent compressed frames.
Each frame that leaves the window is stored as a compact record (\emph{KV Memory} in \Cref{fig:method}), built by \emph{Grouping}, \emph{Factorizing}, and \emph{Summarizing}.
\emph{Cross-Frame Sharing} then lets similar groups in different frames share their statistics (a center, scales, and a low-rank basis) while keeping each frame retrievable.
These steps operate on the key $\mathbf{k}_i$ and value $\mathbf{v}_i\in\mathbb{R}^{D}$ of each token $i$ at every attention layer and KV head, with keys taken before rotary position embedding (RoPE).
Full algorithms are in the supplementary material.

\subsubsection{Grouping}
\label{sec:group}
To encode tokens as groupwise residuals, we partition the $N$ tokens into a set of groups $\mathcal{G}$ in each frame and layer using bisecting $k$-means (BKM) on the joint KV feature $[\bar{\mathbf{k}}_i,\bar{\mathbf{v}}_i]$, where $[\cdot,\cdot]$ denotes concatenation.
We obtain $\bar{\mathbf{k}}_i$ and $\bar{\mathbf{v}}_i$ by stacking keys and values separately across KV heads, centering each set across tokens, and scaling each set to unit root mean square.
BKM starts with one group and proposes binary splits via $k$-means for up to four rounds. 
For each proposed split, we compare the squared error to the assigned group centers before and after splitting.
We accept the split only if the decrease in error per feature dimension exceeds the split threshold $\lambda$.
Once the groups are fixed, all heads share the same group assignments, and each head works on its own $\mathbf{x}_i=[\mathbf{k}_i,\mathbf{v}_i]\in\mathbb{R}^{2D}$.
For each group $g$, the head keeps the center $\boldsymbol{\mu}_g$ and the key and value scales $\boldsymbol{\sigma}_g$. We compute its normalized residual as
$\mathbf{e}_i=(\mathbf{x}_i-\boldsymbol{\mu}_g)\oslash\boldsymbol{\sigma}_g$,
where $\oslash$ denotes element-wise division.

\subsubsection{Factorizing}
\label{sec:factorize}
Grouping similar KV tokens still leaves a $2D$-dimensional residual $\mathbf{e}_i$ per token.
If we fit a low-rank basis directly to each group, each basis vector would still lie in $\mathbb{R}^{2D}$.
To reduce this cost, we project the residuals into a lower-dimensional space shared across groups, then fit each group's basis within that space.
First, for each layer and KV head, we fit a basis $\mathbf{U}\in\mathbb{R}^{2D\times D_U}$ to the residuals from all groups in the first $W$ frames of each video and keep it fixed for later frames.
Each residual is projected into the shared space as $\mathbf{y}_i=\mathbf{U}^\top\mathbf{e}_i$, reducing its dimension from $2D$ to $D_U$.
Second, since each group contains similar KV tokens, we expect much of the variation among its projected residuals to lie along a few group-specific directions. 
We therefore fit a group-specific low-rank basis $\boldsymbol{\Phi}_g\in\mathbb{R}^{D_U\times D_\Phi}$ to the projected residuals $\mathbf{y}_i$ of group $g$.
Each token then stores only its coefficient vector $\mathbf{c}_i=\boldsymbol{\Phi}_g^\top\mathbf{y}_i\in\mathbb{R}^{D_\Phi}$.
We can reconstruct its key and value from $\mathbf{c}_i$ as
\begin{equation}
  \mathbf{x}_i\approx\boldsymbol{\mu}_g+\boldsymbol{\sigma}_g\odot\mathbf{U}\boldsymbol{\Phi}_g\mathbf{c}_i .
  \label{eq:factor}
\end{equation}
Both bases are fitted from the second-moment matrices of $\mathbf{e}_i$ and of $\mathbf{y}_i$ within group $g$, respectively, using pivoted Cholesky.
We use QR decomposition to orthonormalize the selected directions, which are generally not orthogonal.

\subsubsection{Summarizing}
\label{sec:summarize}
\emph{Factorizing} reduces the channels of each token, but a retrieved frame would still be decoded into all $N$ tokens and attended to token by token.
To reduce the tokens read at readout, we keep at most $M$ weighted summary tokens $\{(\mathbf{r}_k,w_k)\}$ instead of all $\{\mathbf{c}_i\}$ for each group.
We cluster $\{\mathbf{c}_i\}$ with $k$-means, take the coefficient vector nearest each cluster center as $\mathbf{r}_k$, and set $w_k$ to the cluster size. For RoPE, we use the spatial position of each selected token.
Selecting a few representatives can change the center and spread of the group's coefficient distribution. 
We update each $\mathbf{r}_k$ with an affine map to align the representatives' weighted mean and covariance with those of the full group.

\subsubsection{Cross-Frame Sharing}
\label{sec:reuse}
When similar groups recur across frames, we reuse previously stored group statistics.
For each group $g$ in the current frame, we first find the recently stored group $p$ whose center is closest to $\boldsymbol{\mu}_g$.
We then find the group in the current frame whose center is closest to $\boldsymbol{\mu}_p$. If that group is $g$, the match is mutual
and $p$ becomes a candidate parent.
Given a candidate parent $p$, we normalize the tokens of $g$ as $\mathbf{e}_i^p=(\mathbf{x}_i-\boldsymbol{\mu}_p) \oslash\boldsymbol{\sigma}_p$ and reconstruct them with $\mathbf{U}\boldsymbol{\Phi}_p$.
For each token, we compute the Euclidean reconstruction errors for keys and values, concatenated across KV heads.
We normalize each error by the root mean square distance of the parent's corresponding key or value vectors from their center, and take the larger normalized error as the token's score.
Group $g$ shares $(\boldsymbol{\mu}_p,\boldsymbol{\sigma}_p, \boldsymbol{\Phi}_p)$ if the mean squared score over its tokens is at most $1$. Without an accepted match, $g$ is factorized as a new parent as in \Cref{sec:factorize}.

\noindent\textbf{Extra residual correction.}
A shared group also stores a single direction $\boldsymbol{\phi}_g$ for what the parent basis misses, namely the direction that captures most of the projected residuals $\mathbf{U}^\top\mathbf{e}_i^{p}$ left outside the span of $\boldsymbol{\Phi}_p$.
Each summary token stores one scalar coefficient $\eta_k$ along $\boldsymbol{\phi}_g$, set to match the moments of the summarized tokens, and is decoded as
\begin{equation}
  \hat{\mathbf{x}}_k=\boldsymbol{\mu}_p+\boldsymbol{\sigma}_p\odot\mathbf{U}\bigl(\boldsymbol{\Phi}_p\mathbf{r}_k+\boldsymbol{\phi}_g\eta_k\bigr),
  \label{eq:reuse}
\end{equation}
where a new parent has $p{=}g$ and $\eta_k{=}0$.

\noindent\textbf{Memory.}
For each video, \ours stores one basis $\mathbf{U}$ per KV head in each layer.
A new parent stores its own statistics $(\boldsymbol{\mu}_g,\boldsymbol{\sigma}_g,\boldsymbol{\Phi}_g)$. A shared group reuses statistics from its parent and stores an additional residual direction $\boldsymbol{\phi}_g$ and coefficients $\{\eta_k\}$.
\Cref{fig:method} shows the shared basis and reusable group statistics in green and the compact records of past frames in orange.
Each record has its own summary tokens, coefficients, and positions, so each frame can be retrieved individually.
Records default to FP16, with optional INT8 group statistics for further memory savings.

\subsection{Look-Back Gate}
\label{sec:gate}
The gate decides whether a question $q$ is answered from the recent frames only or by looking back at past frames.
If recent frames contain evidence relevant to $q$, attending to them should change how the model represents $q$.
To measure this change, we process two copies of $q$ and its answer choices in a single forward pass, masking access to the recent frames for one copy.
We extract the last-token states $\mathbf{h}^{\mathrm{rec}}_l(q)$ and $\mathbf{h}^{\mathrm{txt}}_l(q)$ from the two copies at layers $l\in\mathcal{L}$.
The gate score $\delta(q)$ is the cosine distance $1-\cos(\mathbf{h}^{\mathrm{rec}}_l(q),\mathbf{h}^{\mathrm{txt}}_l(q))$ averaged over $\mathcal{L}$.
We interpret a small $\delta(q)$ as a sign that the recent frames may provide insufficient evidence to answer $q$.
The gate therefore retrieves past frames when $\delta(q)\le\tau$ and uses only the recent frames otherwise without any training.

\subsection{Retrieval and Readout}
\label{sec:readout}
On the look-back route, \ours scores each past frame by comparing its retrieval key with the question's attention queries.
As in ReKV~\cite{rekv2025}, this key is the mean of the frame's pre-RoPE keys before compression, kept per layer and KV head.
The highest-scoring records are retrieved beside the recent window (\emph{Score \& retrieve} in \Cref{fig:method}).
We decode each summary token into $\hat{\mathbf{x}}_k=[\hat{\mathbf{k}}_k,\hat{\mathbf{v}}_k]$ with \Cref{eq:reuse}.
Let $\tilde{\mathbf{k}}_k$ denote the key obtained by applying RoPE to $\hat{\mathbf{k}}_k$ using the frame's position in the current context and the summary token's stored spatial position (\Cref{sec:summarize}).
For an attention query $\mathbf{q}$, the summary token then has the logit $\mathbf{q}^\top\tilde{\mathbf{k}}_k/\sqrt{D}+\log w_k$, so that it counts as $w_k$ tokens in the softmax, as in the proportional attention of ToMe~\cite{tome2023}.

\vspace{-2mm}

\section{Experiments}
\label{sec:exp}

\subsection{Experiment Setup}
\label{sec:setup}

\noindent\textbf{Benchmarks.}
We evaluate on \data, which contains 1,320 questions across 440 distinct videos. We compute question-level accuracy over all questions and paired accuracy for \emph{Cross-time} and \emph{Cross-state} over 440 pairs each (see the supplementary material for detailed statistics).
We also evaluate on 1,468 OVO-Bench~\cite{ovobench2025} questions (837 real-time and 631 backward), the 2,500 real-time questions from StreamingBench~\cite{streamingbench2026}, and all 4,608 RTV-Bench~\cite{rtvbench2025} questions.

\noindent\textbf{Implementation details.} We use LLaVA-OneVision-7B (LLaVA-OV)~\cite{llavaonevision2025} and Qwen2.5-VL-7B~\cite{qwen25vl2025} at 1\,fps on NVIDIA RTX A6000 GPUs.
We re-run each training-free baseline using its public code and released settings.
In the tables, dashed lines separate the recent-frame, bounded-memory, and retrieval-based baselines, in that order.
For LLaVA-OV, the recent window holds $W{=}10$ frames, and we set $D_U{=}48$, $D_\Phi{=}4$, $M{=}8$, and $\lambda{=}2.5$.
BKM runs for at most four rounds (${\le}16$ groups).
For cross-frame sharing, we search the groups from the previous frame and the 256 most recently stored groups.
The gate and the recent-only route both use the last five frames.
The gate uses layers 10--18 of the 28-layer backbone with a default threshold of $\tau{=}0.015$.
The look-back route retrieves up to 54 past frames.
For \data{}, we use ``Near the beginning of this clip'' as the default prefix for \Ptwo{}.
All reported memory includes metadata such as retrieval indexes and excludes the recent window.
Detailed Qwen2.5-VL settings and hyperparameter sensitivity are in the supplementary material.

\begin{table}[t]
\centering
\small
\setlength{\tabcolsep}{1.35pt}
\caption{\textbf{Accuracy (\%) on \data.} $s@t$: question about $s$ asked at $t$. \emph{Cross-time} (\Cone, \Ptwo); \emph{Cross-state} (\Ptwo, \Ctwo), \emph{Consistency}: $s_2@t_2$ accuracy conditioned on \emph{Cross-time} correct cases. Best in bold, second-best underlined.}
\label{tab:sce}
\begin{tabular}{l ccc c c c}
\toprule
\multirow{2}{*}[-1.5pt]{\hd{Method}} & \hd{$@t_1$} & \multicolumn{2}{c}{\hd{$@t_2$}} & \multicolumn{2}{c}{\hd{Cross-}} & \multirow{2}{*}[-1.5pt]{\hd{Consistency}} \\[-1pt]
\cmidrule(lr){2-2}\cmidrule(lr){3-4}\cmidrule(lr){5-6} & \hd{$s_1$} & \hd{$s_1$} & \hd{$s_2$} & \hd{time} & \hd{state} & \\
\midrule
\rowcolor{bbrow}\textbf{LLaVA-OV-7B}~\cite{llavaonevision2025} &  &  &  &  &  &  \\
\rf{\quad Text only (no video)} & \rf{28.9} & \rf{31.8} & \rf{29.1} & \rf{22.0} & \rf{7.0} & \rf{18.6} \\
\famsep{7}
\rep{\plus SimpleStream~\cite{simplestream2026}} & \rep{54.3} & \rep{23.9} & \rep{50.5} & \rep{16.8} & \rep{4.8} & \rep{21.6} \\
\famsep{7}
\rep{\plus HERMES~\cite{hermes2026}} & \rep{57.7} & \rep{27.5} & \rep{45.9} & \rep{21.4} & \rep{7.0} & \rep{26.6} \\
\rep{\plus CausalMem~\cite{causalmem2026}} & \rep{53.4} & \rep{32.0} & \rep{41.6} & \rep{25.7} & \rep{10.0} & \rep{33.6} \\
\rep{\plus CoRDS~\cite{cords2026}} & \rep{57.3} & \rep{34.8} & \rep{44.5} & \rep{28.0} & \rep{11.1} & \rep{37.4} \\
\rep{\plus InfiniPot-V~\cite{infinipotv2025}} & \rep{53.2} & \rep{38.6} & \rep{41.8} & \rep{30.7} & \rep{13.2} & \rep{35.6} \\
\famsep{7}
\plus ReKV~\cite{rekv2025} & 57.3 & 48.4 & 33.2 & \textbf{40.2} & 14.8 & 30.5 \\
\plus StreamKV~\cite{streamkv2026} & 55.5 & 46.1 & 33.2 & 38.4 & 15.2 & 33.1 \\
\plus MuKV~\cite{mukv2026} & 52.7 & 36.8 & 30.5 & 28.2 & 7.5 & 20.2 \\
\plus StreamingTOM~\cite{streamingtom2026} & 52.5 & 43.6 & 34.3 & 34.1 & 11.6 & 27.3 \\
\plus ShallowStream~\cite{shallowstream2026} & 58.9 & 30.7 & 51.4 & 22.0 & 7.5 & 22.7 \\
\rowcolor{oursrow}\plus \ours & 56.6 & 47.3 & 47.5 & \underline{39.1} & \underline{22.0} & \underline{49.4} \\
\rowcolor{oursrow}\plus \ours (INT8) & 56.6 & 47.5 & 47.5 & \underline{39.1} & \textbf{22.3} & \textbf{50.0} \\
\midrule
\rowcolor{bbrow}\textbf{Qwen2.5-VL-7B}~\cite{qwen25vl2025} &  &  &  &  &  &  \\
\rep{\plus SimpleStream~\cite{simplestream2026}} & \rep{60.0} & \rep{30.7} & \rep{46.8} & \rep{24.1} & \rep{3.6} & \rep{11.3} \\
\famsep{7}
\rep{\plus HERMES~\cite{hermes2026}} & \rep{60.5} & \rep{50.0} & \rep{38.0} & \rep{42.3} & \rep{15.0} & \rep{30.6} \\
\famsep{7}
\plus ReKV~\cite{rekv2025} & 60.2 & 48.2 & 37.0 & 42.3 & 15.9 & 32.3 \\
\rowcolor{oursrow}\plus \ours & 59.8 & 51.4 & 40.7 & \textbf{43.9} & \textbf{17.7} & \underline{35.8} \\
\rowcolor{oursrow}\plus \ours (INT8) & 59.8 & 50.9 & 40.7 & \underline{43.6} & \underline{17.5} & \textbf{35.9} \\
\bottomrule
\end{tabular}
\end{table}

\noindent\textbf{Metrics.} We report individual accuracy for all benchmarks and paired accuracy for \data.
\emph{Cross-time} pairs \Cone with \Ptwo.
Both questions refer to $s_1$, so this metric assesses whether the model can identify a state when it is current and recall it after the state changes.
\emph{Cross-state} pairs \Ptwo with \Ctwo.
The two questions share the same video history and refer to the prior state $s_1$ and the current state $s_2$.
This metric assesses whether the model can distinguish the two states based on the temporal reference in each question.
We also report \emph{Consistency}, an auxiliary metric similar to consistency in VQA~\cite{hudson2019gqa,selvaraju2020squinting}.
It measures the accuracy of \Ctwo on \emph{Cross-time} correct cases, indicating whether recalling a prior state comes with tracking current states.

\begin{table}[t]
\centering
\small
\setlength{\tabcolsep}{3.6pt}
\caption{\textbf{Accuracy (\%) on existing benchmarks.} OVO: mean of the real-time (RT) and backward (BW) averages; SB: StreamingBench real-time split (ten tasks); RTV: overall.}
\label{tab:public}
\begin{tabular}{l ccc c c}
\toprule
\multirow{2}{*}[-4.5pt]{\hd{Method}} & \multicolumn{3}{c}{\hd{OVO-Bench}} & \hd{SB} & \hd{RTV} \\[2.6pt]
\cmidrule(lr){2-4} & \hd{RT} & \hd{BW} & \hd{Overall} & & \\[2.6pt]
\midrule
\rowcolor{bbrow}\textbf{LLaVA-OV-7B}~\cite{llavaonevision2025} & & & & & \\
\plus SimpleStream~\cite{simplestream2026} & \underline{72.8} & 44.8 & \underline{58.8} & 72.1 & \underline{55.8} \\
\famsep{6}
\plus HERMES~\cite{hermes2026} & 69.9 & \underline{50.3} & \textbf{60.1} & \textbf{74.2} & 52.9 \\
\plus CausalMem~\cite{causalmem2026} & 66.2 & 44.3 & 55.3 & 71.3 & 50.8 \\
\plus CoRDS~\cite{cords2026} & 67.1 & \textbf{50.5} & \underline{58.8} & \underline{73.4} & 47.7 \\
\plus InfiniPot-V~\cite{infinipotv2025} & 68.0 & 45.2 & 56.6 & 71.8 & 51.4 \\
\famsep{6}
\plus ReKV~\cite{rekv2025} & 63.1 & 46.1 & 54.6 & 69.2 & 47.9 \\
\plus StreamKV~\cite{streamkv2026} & 64.4 & 43.3 & 53.8 & 70.4 & 48.4 \\
\plus MuKV~\cite{mukv2026} & 51.5 & 42.5 & 47.0 & 60.5 & 42.8 \\
\plus StreamingTOM~\cite{streamingtom2026} & 60.3 & 42.2 & 51.2 & 68.3 & 46.7 \\
\plus ShallowStream~\cite{shallowstream2026} & \textbf{73.2} & 44.2 & 58.7 & 70.6 & \textbf{56.8} \\
\rowcolor{oursrow}\plus \ours & 71.4 & 48.8 & \textbf{60.1} & 73.2 & 53.5 \\
\rowcolor{oursrow}\plus \ours (INT8) & 71.3 & 49.0 & \textbf{60.1} & 73.2 & 53.5 \\
\midrule
\rowcolor{bbrow}\textbf{Qwen2.5-VL-7B}~\cite{qwen25vl2025} & & & & & \\
\plus SimpleStream~\cite{simplestream2026} & \textbf{68.6} & \underline{47.6} & 58.1 & 71.3 & \textbf{56.0} \\
\famsep{6}
\plus HERMES~\cite{hermes2026} & 64.7 & 46.3 & 55.5 & \textbf{74.2} & 53.8 \\
\famsep{6}
\plus ReKV~\cite{rekv2025} & 65.7 & 44.5 & 55.1 & 73.0 & 51.2 \\
\rowcolor{oursrow}\plus \ours & 67.3 & \textbf{50.1} & \underline{58.7} & \underline{73.4} & \underline{54.4} \\
\rowcolor{oursrow}\plus \ours (INT8) & \underline{67.4} & \textbf{50.1} & \textbf{58.8} & \underline{73.4} & \underline{54.4} \\
\bottomrule
\end{tabular}
\end{table}

\subsection{Main Results}
\label{sec:main}
\begin{figure}[t]
    \centering
    \includegraphics[width=0.95\linewidth]{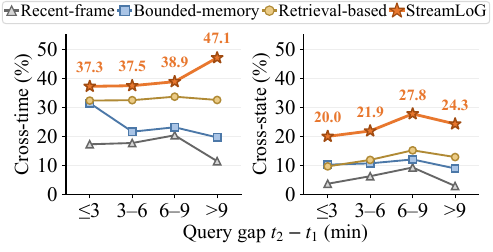}
    \vspace{-1mm}
    \caption{\textbf{\emph{Cross-time} and \emph{Cross-state} accuracy by query gap on SCE.} Recent-frame shows SimpleStream; bounded-memory and retrieval-based curves show family averages.}
    \label{fig:gap}
\end{figure}

\noindent\textbf{\data.}
On LLaVA-OV, \Cref{tab:sce} shows a trade-off between current-state tracking and prior-state recall. Bounded-memory methods generally score higher on current-state questions (\Cone and \Ctwo) than on prior-state questions (\Ptwo), and their low \Ptwo accuracy limits both paired metrics.
Retrieval-based methods generally score higher on \Ptwo and \emph{Cross-time}, but their lower \Ctwo scores indicate difficulty identifying the current state at $t_2$, and their \emph{Cross-state} scores remain low.
Both precision variants of our method lead on \emph{Cross-state} and \emph{Consistency} across both architectures, with respective FP16 gains
of 6.8 and 12.0 percentage points over the best methods on LLaVA-OV
while retaining near-best \emph{Cross-time} accuracy.
This pattern suggests that \ours recalls prior states without losing track of the current state.
When we break down the results by query gap, our method stays above the family average curves in every bin for \emph{Cross-time} and \emph{Cross-state} (\Cref{fig:gap}).
Even for gaps longer than 9\,min, \ours achieves 47.1\% on \emph{Cross-time} and 24.3\% on \emph{Cross-state}.

\noindent\textbf{Existing streaming benchmarks.}
\ours remains competitive across all three benchmarks on both backbones (\Cref{tab:public}, with per-task results in the supplementary material).
Notably, SimpleStream~\cite{simplestream2026} is also competitive with only the five most recent frames. 
On LLaVA-OV, it ties for second overall on OVO-Bench and ranks second on RTV-Bench.
Its OVO-Bench backward-tracing score of 44.8\% is close to most baselines including both bounded-memory and retrieval-based methods. 
On \data, however, SimpleStream answers both \emph{Cross-state} questions correctly in only 4.8\% of pairs (see \Cref{tab:sce}).
This contrast suggests that strong performance on existing benchmarks offers limited evidence that a model can recall a prior state and distinguish it from the current one after a change.

\section{In-Depth Analysis}
\label{sec:analysis}

\begin{table}[t]
\centering
\small
\setlength{\tabcolsep}{1.25pt}
\caption{\textbf{Component ablation.} Fact.: Factorize and Summarize with one group per frame; Grp.: BKM groups per frame; Share: cross-frame sharing with residual correction. Mem.: \% of raw KV.}
\label{tab:design}
\begin{tabular}{*{4}{W{c}{\dimexpr(\columnwidth-18\tabcolsep-30pt)/8\relax}} W{c}{30pt} *{4}{W{c}{\dimexpr(\columnwidth-18\tabcolsep-30pt)/8\relax}}}
\toprule
\multicolumn{4}{c}{\hd{Component}} & \hd{Mem.$\downarrow$}
  & \multicolumn{2}{c}{\hd{$@t_2$}}
  & \multicolumn{2}{c}{\hd{Cross-}} \\[-1pt]
\cmidrule(lr){1-4}\cmidrule(lr){6-7}\cmidrule(lr){8-9}
\hd{Fact.} & \hd{Grp.} & \hd{Share} & \hd{Gate}
  & & \hd{$s_1$} & \hd{$s_2$} & \hd{time} & \hd{state} \\
\midrule
    &     &     &     & 100.0 & 47.7 & 36.4 & \textbf{40.2} & 16.1 \\
\cm &     &     &     & \textbf{1.3} & 35.9 & 46.8 & 28.6 & 12.7 \\
\cm & \cm &     &     & 11.8 & 47.3 & 40.7 & 39.3 & 17.5 \\
\cm & \cm & \cm &     & \underline{5.8} & 48.0 & 42.0 & \underline{39.8} & \underline{18.4} \\
\rowcolor{oursrow}
\cm & \cm & \cm & \cm & \underline{5.8} & 47.3 & 47.5 & 39.1 & \textbf{22.0} \\
\bottomrule
\end{tabular}
\end{table}

\subsection{Component Ablation}
\label{sec:ablation}
\Cref{tab:design} compares compression and routing on \data using uncompressed KV with always-on retrieval as the reference.
Under always-on retrieval, compressing each frame as a single group reduces prior-state recall ($s_1@t_2$) while improving current-state accuracy ($s_2@t_2$), suggesting that retrieved past evidence can interfere with current-state predictions.
BKM grouping recovers most of the lost prior-state recall and partially reverses the gain in current-state accuracy.
Cross-frame sharing roughly halves the storage of the grouped representation from $11.8\%$ to $5.8\%$ of raw KV with comparable accuracy.
Adding the gate to compressed KV with cross-frame sharing raises $s_2@t_2$ by $5.5$ percentage points, with little change in $s_1@t_2$.
This pattern suggests that the look-back gate mitigates interference by retrieving past frames selectively.
Together, these results highlight the value of pairing a compact memory that largely preserves prior-state recall with selective retrieval that improves current-state accuracy.

\subsection{Comparison of Memory Designs}
\label{sec:ana_group}
The first row of \Cref{tab:repr} gives uncompressed KV as the reference.
Storing group centers alone lowers prior-state accuracy from $47.7\%$ to $44.5\%$.
Adding full-dimensional residuals recovers most of this loss, showing the value of retaining within-group variation.
Using the shared basis $\mathbf{U}$ alone gives lower current-state accuracy than fitting a separate basis $\mathbf{\Phi_g}$ to each group. 
Since $\mathbf{U}$ is fitted on the first $W$ frames and kept fixed, it may miss residual directions that matter for later groups.
A group-specific basis $\Phi_g$ can adapt to these directions but direct factorization requires storing a separate set of basis vectors in $\mathbb{R}^{2D}$ for each group.
The two-stage representation retains comparable prior- and current-state accuracy at 11.9\% of raw KV storage, less than half that of either single-stage variant. We also vary only the KV compression strategy while keeping the rest of our pipeline fixed (\Cref{tab:compression_swap}).
At comparable mean memory, \ours outperforms the other compressors on prior-state recall and both paired metrics, highlighting the benefit of sharing recurring KV statistics across frames while retaining frame-specific summaries.

\begin{table}[t]
\centering
\small
\setlength{\tabcolsep}{1.25pt}
\caption{\textbf{Effect of compression design} (always look back, no summarizing or cross-frame sharing). $\boldsymbol{\mu}_g$: group center; $\mathbf{e}_i$: residuals; $\mathbf{U}$, $\boldsymbol{\Phi}_g$: shared and groupwise bases. Mem.: \% of raw KV.}
\label{tab:repr}
\begin{tabular}{*{4}{W{c}{\dimexpr(\columnwidth-18\tabcolsep-30pt)/8\relax}} W{c}{30pt} *{4}{W{c}{\dimexpr(\columnwidth-18\tabcolsep-30pt)/8\relax}}}
\toprule
\multicolumn{4}{c}{\hd{Component}} & \hd{Mem.$\downarrow$}
  & \multicolumn{2}{c}{\hd{$@t_2$}}
  & \multicolumn{2}{c}{\hd{Cross-}} \\[-1pt]
\cmidrule(lr){1-4}\cmidrule(lr){6-7}\cmidrule(lr){8-9}
\hd{$\boldsymbol{\mu}_g$} & \hd{$\mathbf{e}_i$} & \hd{$\mathbf{U}$}
  & \hd{$\boldsymbol{\Phi}_g$} & & \hd{$s_1$} & \hd{$s_2$}
  & \hd{time} & \hd{state} \\
\midrule
    &     &     &     & 100.0 & 47.7 & 36.4 & \textbf{40.2} & 16.1 \\
\cm &     &     &     & \textbf{6.0} & 44.5 & 40.5 & 36.1 & 14.3 \\
\cm & \cm &     &     & 105.8 & 47.5 & 36.4 & \underline{40.0} & 16.1 \\
\cm & \cm & \cm &     & 24.8 & 47.0 & 37.7 & 39.8 & 16.6 \\
\cm & \cm &     & \cm & 30.3 & 46.6 & 40.5 & 38.9 & \textbf{18.6} \\
\rowcolor{oursrow}
\cm & \cm & \cm & \cm & \underline{11.9} & 47.5 & 39.8 & 39.3 & \underline{17.3} \\
\bottomrule
\end{tabular}
\end{table}

\begin{table}[t]
\centering
\small
\setlength{\tabcolsep}{1.25pt}
\caption{\textbf{KV compression strategies on \data.} Only the compressor in \ours is replaced, at approximately matched memory. Mem.: \% of raw KV.}
\label{tab:compression_swap}
\begin{tabular}{W{l}{\dimexpr\columnwidth-12\tabcolsep-140pt\relax}
                W{c}{30pt} *{4}{W{c}{27.5pt}}}
\toprule
\multirow{2}{*}[-1.5pt]{%
  \shortstack[l]{\hd{Compression}\\\hd{Strategy}}}
& \multirow{2}{*}[-1.5pt]{\hd{Mem.$\downarrow$}}
& \multicolumn{2}{c}{\hd{$@t_2$}}
& \multicolumn{2}{c}{\hd{Cross-}} \\[-1pt]
\cmidrule(lr){3-4}\cmidrule(lr){5-6}
& & \hd{$s_1$} & \hd{$s_2$} & \hd{time} & \hd{state} \\
\midrule
None
& 100.0 & 47.3 & 43.0
& \textbf{39.5} & \underline{19.3} \\
StreamKV~\cite{streamkv2026}
& 5.8 & 30.7 & 50.9
& 22.7 & 10.5 \\
MuKV~\cite{mukv2026}
& 5.8 & 35.2 & 50.2
& 28.0 & 14.8 \\
StreamingTOM~\cite{streamingtom2026}
& 5.9 & 41.1 & 48.2
& 34.3 & 17.7 \\
\rowcolor{oursrow}\ours~\textbf{(Ours)}
& 5.8 & 47.3 & 47.5
& \underline{39.1} & \textbf{22.0} \\
\bottomrule
\end{tabular}
\end{table}

\begin{table}[t]
\centering
\small
\setlength{\tabcolsep}{1pt}
\caption{\textbf{Comparison of look-back rules.} Random: random routing at our gate's rate. Text: question-only routing. Oracle: picks whichever of our two routes answers correctly. LB: look-back rate (\%) averaged over four benchmarks.}
\label{tab:route}
\begin{tabularx}{\columnwidth}{l *{4}{Y} cccc}
\toprule
\hd{Rule} & \multicolumn{4}{c}{\bfseries \data} & \hd{OVO} & \hd{SB} & \hd{RTV} & \hd{LB} \\
\cmidrule(lr){2-5}
 & \multicolumn{2}{c}{\hd{$@t_2$}} & \multicolumn{2}{c}{\hd{Cross-}} & & & & \\[-1pt]
\cmidrule(lr){2-3}\cmidrule(lr){4-5}
 & \hd{$s_1$} & \hd{$s_2$} & \hd{time} & \hd{state} & & & & \\
\midrule
None & 30.0 & 50.9 & 21.4 & 8.4 & \textbf{61.4} & 71.9 & \textbf{55.4} & -- \\
Always & 48.0 & 42.0 & \textbf{39.8} & \underline{18.4} & 57.2 & \textbf{73.4} & 52.0 & 100.0 \\
Random & 41.6 & 45.2 & 33.3 & 16.1 & 58.9 & 72.9 & \underline{53.5} & 60.9 \\
Text~\cite{shallowstream2026} & 30.7 & 51.1 & 21.6 & 9.5 & \underline{60.3} & 71.8 & \textbf{55.4} & 3.9 \\
\rowcolor{oursrow}\ours~\textbf{(Ours)} & 47.3 & 47.5 & \underline{39.1} & \textbf{22.0} & 60.1 & \underline{73.2} & \underline{53.5} & 60.9 \\
\midrule
\rf{Oracle} & \rf{56.1} & \rf{56.4} & \rf{45.9} & \rf{28.9} & \rf{67.9} & \rf{78.6} & \rf{60.9} & \rf{--} \\
\bottomrule
\end{tabularx}
\end{table}

\subsection{Analysis of the Look-Back Gate}
\label{sec:ana_gate}
We fix $\tau{=}0.015$ for all benchmarks (sweeps over $\tau$ in the supplementary material) and compare look-back rules with our memory in \Cref{tab:route}.
Question-only routing~\cite{shallowstream2026} rarely triggers look-back (3.9\% on average across the four benchmarks) and achieves scores close to those of answering from recent frames alone (\emph{None} in \Cref{tab:route}).
On \data, its scores are close to those of always answering from recent frames.
Our gate nearly matches the prior-state accuracy ($s_1@t_2$) of always looking back while recovering most of its current-state ($s_2@t_2$) loss, achieving the best \emph{Cross-state} score.
Random routing at the same rate falls behind on every \data metric, so the gain comes from which questions look back, not from how often.
On OVO-Bench and RTV-Bench, the gate outperforms always-on retrieval but falls short of recent-only answering.
The oracle analysis points out an opportunity in avoiding retrieval that corrupts recent-only answers (see the supplementary material).

\subsection{Memory and Latency}
\label{sec:ana_mem}
\begin{figure}[t]
  \centering
  \begin{subfigure}[b]{0.543\linewidth}\centering
    \includegraphics[width=\linewidth,trim={0 2pt 0 0},clip]{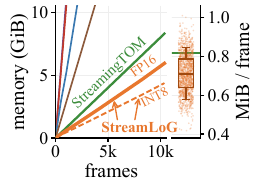}
    \caption{Memory size}\label{fig:eff_mem}
  \end{subfigure}\hfill
  \begin{subfigure}[b]{0.457\linewidth}\centering
    \includegraphics[width=\linewidth,trim={0 2pt 0 0},clip]{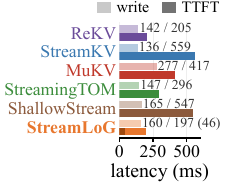}
    \caption{Latency}\label{fig:eff_speed}
  \end{subfigure}
    \caption{\textbf{Memory and latency comparison.} (a) Left: memory size for a 10,360-frame video. Right: distribution of our FP16 memory per frame (whiskers: 10th--90th percentiles; line: StreamingTOM). (b) Per-frame write time and time to first token (TTFT). Darker segment and parentheses: recent-only route.}
  \label{fig:eff}
  \vspace{-2.5mm}
\end{figure}

\noindent\textbf{Memory.}
Over the 1,323 streams of the existing benchmarks, our FP16 memory uses about $1/16$ of raw KV storage per frame.
On the 10,360-frame video, our memory grows more slowly than that of the baselines (\Cref{fig:eff_mem}, left).
For most videos, our method also uses less memory per frame than StreamingTOM (right).
As low-bit storage is orthogonal to dimension reduction, INT8 further lowers memory with little loss in accuracy (\Cref{tab:sce,tab:public}).

\noindent\textbf{Latency.}
On LLaVA-OV, we decode at most $|\mathcal{G}|M$ summary tokens per retrieved frame, averaging about 90 instead of 196 visual tokens.
On a 1,024-frame stream, our per-frame write time adds a modest overhead over ReKV~\cite{rekv2025} while look-back TTFT remains comparable (\Cref{fig:eff_speed}; settings and per-stage timings in the supplementary material).
The recent-only route skips retrieval and decoding of past KV records, reducing TTFT to 46 ms.

\subsection{Robustness of SCE Construction}
\label{sec:sensitivity}
\begin{figure}[t]
    \centering
    \includegraphics[width=0.97\linewidth]{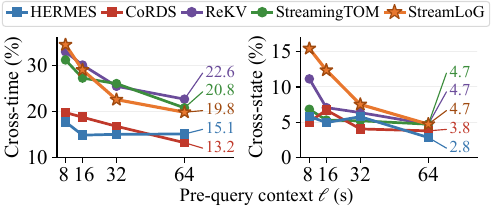}
    \caption{\textbf{Pre-query context length sensitivity.} \emph{Cross-time} and \emph{Cross-state} accuracy under different pre-query context lengths.}
    \label{fig:precontext}
\end{figure}

\noindent\textbf{Pre-query context length.} 
The default $8$\,s pre-query context may truncate earlier evidence of $s_1$ and affect the relative performance of methods.
We examine whether the \emph{Cross-time} and \emph{Cross-state} advantage of retrieval-based methods persists with longer pre-query contexts.
We rebuild clips with $\ell\in\{8,16,32,64\}$\,s and compare \ours with two bounded-memory baselines~\cite{hermes2026,cords2026} and two retrieval-based baselines~\cite{rekv2025,streamingtom2026}.
At each length, we retain cases in which $t_1$ falls within the first quarter of the clip. Because $t_1$ can fall near the end of that quarter, we use ``Earlier in the clip'' as the \Ptwo{} prefix at all four lengths.
At each evaluated length, retrieval-based baselines achieve higher \emph{Cross-time} accuracy than the bounded-memory baselines (\Cref{fig:precontext}).
These results are consistent with a benefit from accessing earlier frames for prior-state recall.
Our \emph{Cross-state} gains also extend beyond the default length, although the gap narrows as the context grows.

\noindent\textbf{Prefix sensitivity.} Since the prefix specifies which state is queried, its wording could favor some methods.
We replace ``Near the beginning of this clip'' with four alternatives, such as ``At the start of this clip'' (full list in the supplementary material).
Over the five wordings (\Cref{tab:prefix}), our worst \emph{Cross-state} and \emph{Consistency} scores exceed every baseline's mean, so our gains do not depend on a particular wording.

\begin{table}[t]
{\centering\small
\setlength{\tabcolsep}{2.2pt}
\caption{\textbf{Prefix sensitivity.} \emph{Cross-time}, \emph{Cross-state}, and \emph{Consistency} (\%) on \data with the original prefix and four alternatives.}
\label{tab:prefix}
\begin{tabular}{l cc cc cc}
\toprule
\multirow{2}{*}{\hd{Method}} & \multicolumn{2}{c}{\hd{Cross-time}} & \multicolumn{2}{c}{\hd{Cross-state}} & \multicolumn{2}{c}{\hd{Consistency}} \\
\cmidrule(lr){2-3}\cmidrule(lr){4-5}\cmidrule(lr){6-7}
 & \hd{Mean} & \hd{Worst} & \hd{Mean} & \hd{Worst} & \hd{Mean} & \hd{Worst} \\
\midrule
HERMES~\cite{hermes2026} & 21.3 & 20.9 & 7.7 & 6.8 & 29.0 & 25.8 \\
CoRDS~\cite{cords2026} & 27.1 & 26.4 & 11.1 & 10.0 & \underline{37.9} & \underline{35.3} \\
\famsep{7}
ReKV~\cite{rekv2025} & \textbf{39.9} & \textbf{39.1} & \underline{15.3} & \underline{14.8} & 31.8 & 30.5 \\
StreamingTOM~\cite{streamingtom2026} & 34.3 & 33.6 & 11.5 & 10.9 & 27.6 & 27.0 \\
\famsep{7}
\rowcolor{oursrow}\ours\textbf{~(Ours)} & \underline{37.8} & \underline{36.1} & \textbf{20.3} & \textbf{18.6} & \textbf{48.2} & \textbf{46.5} \\
\bottomrule
\end{tabular}
\par}
\end{table}

\section{Conclusion}
\label{sec:conc}
We revisited compact streaming video memory with \data, whose paired questions assess whether a model can identify a state when it is current and recall it after the state changes (\emph{Cross-time}), and distinguish it from the current state (\emph{Cross-state}).
We found that existing methods often favor either current-state tracking or prior-state recall, and that retrieved prior evidence can interfere with current-state answers.
To preserve both, we introduced \ours, a training-free compact memory with a look-back gate.
The memory compresses past frames with shared and groupwise low-rank residuals, keeping each frame retrievable.
The gate retrieves only when recent frames appear insufficient.
At about $1/16$ of raw KV storage per frame, we achieve strong paired scores and remain competitive on existing benchmarks.
Building on the initial findings from \data, future work could develop a standalone benchmark with annotated state changes to assess the ability to retain and distinguish multiple states over time.

\clearpage
\section*{LLM Usage Statement}
\label{sec:llm}
We used Claude (Anthropic) and ChatGPT (OpenAI) to proofread author-written text. Under our direction, these tools assisted with method implementation, automation of sequential experiment runs, result aggregation, and plotting. The authors developed the research ideas and designed the experiments without LLM assistance. The authors also verified the citations, checked all plotted values against the underlying evaluation results, and interpreted the findings. The authors take full responsibility for the manuscript and its reported results.

{
    \small
    \bibliographystyle{ieeenat_fullname}
    \bibliography{main}
}

\algrenewcommand\algorithmicrequire{\textbf{Input:}}
\algrenewcommand\algorithmicensure{\textbf{Output:}}
\algblockdefx{Stage}{EndStage}[1]{\textbf{#1}}{}
\algnotext{EndStage}
\newcommand{\Mod}[1]{\textbf{#1}}
\newcommand{\AlgLines}[1]{\begin{tabular}[t]{@{}l@{}}#1\end{tabular}}
\providecommand{\cmt}{}\renewcommand{\cmt}[1]{{\color{blue}#1}}
\renewcommand{\thesection}{\Alph{section}}
\renewcommand{\thetable}{\Alph{table}}
\renewcommand{\thefigure}{\Alph{figure}}
\renewcommand{\thealgorithm}{\Alph{algorithm}}
\AtBeginDocument{\crefname{algorithm}{Alg.}{Algs.}\Crefname{algorithm}{Algorithm}{Algorithms}%
  \crefname{section}{Appendix}{Appendices}\Crefname{section}{Appendix}{Appendices}%
  \crefname{subsection}{Appendix}{Appendices}\Crefname{subsection}{Appendix}{Appendices}}
\renewcommand{\dbltopfraction}{0.95}
\renewcommand{\textfraction}{0.05}
\setcounter{dbltopnumber}{2}
\makeatletter
\renewcommand\thesubtable{(\alph{subtable})}
\renewcommand\p@subtable{\thetable}
\makeatother
\captionsetup[subtable]{labelformat=simple}

\appendix
\renewcommand\thefigure{\Alph{figure}}    
\setcounter{figure}{0}  
\renewcommand\thetable{\Alph{table}}
\setcounter{table}{0} 
\newcommand{\midsize}{\fontsize{8.5}{10.2}\selectfont} %

\maketitlesupplementary
\suppressfloats[t] 

This supplementary material provides the details of \data and \ours that the main paper refers to, together with further analyses and the per-task results on each existing benchmark. It is organized as follows:
\begin{itemize}[leftmargin=*,itemsep=1pt,topsep=2pt]
\item \textbf{\Cref{sec:s_sce}}: construction and statistics of \data, results by source and question type, and robustness to the answer-choice order, the pre-query context length, the prefix, and smaller backbones.
\item \textbf{\Cref{sec:s_method}}: algorithms and equations of the \ours memory, the look-back gate, and the readout.
\item \textbf{\Cref{sec:s_impl}}: settings of the backbones and baselines.
\item \textbf{\Cref{sec:s_analysis}}: sensitivity to hyperparameters, choice of grouping, residual correction, and the look-back gate.
\item \textbf{\Cref{sec:s_mem}}: per-stage write time and time to first token (TTFT) of ours and the retrieval-based baselines.
\item \textbf{\Cref{sec:s_pertask}}: per-task results on the existing benchmarks.
\end{itemize}

\section{SCE Details}
\label{sec:s_sce}

\subsection{Construction and Statistics}
\label{sec:s_construct}
\label{sec:s_stats}
\noindent\textbf{Construction.} Each case of \data uses a video and question from RTV-Bench~\cite{rtvbench2025}, StreamingBench~\cite{streamingbench2026}, or OVO-Bench~\cite{ovobench2025}. The answer to the question changes from $s_1$ at query time $t_1$ to $s_2$ at a later query time $t_2$. Matching the exact question text within each video gives 1,957 candidate pairs, each with at least 5\,s between $t_1$ and $t_2$, at least 8\,s of video before $t_1$, and enough answer choices in its two source questions to form four distinct choices including $s_1$ and $s_2$. We keep the pair with the widest gap $t_2-t_1$ in each video and manually review the retained pairs, yielding 440 cases. We also manually remove phrases that only restate the query time (\eg, ``just now'', ``right now'', ``at this moment''), since each question already refers to the state at the time it is asked. We keep the rest of each source question unchanged. For example, ``so far'' stays in counting questions because it defines what is counted. \Cone and \Ctwo keep the source answer choices. For \Ptwo, we reuse the answer choices of \Cone and replace a wrong choice with the distractor $s_2$ if absent.

\begin{table}[t]
\centering
\footnotesize
\caption{\textbf{Statistics of \data by source.} (a) Cases, median query gap $t_2-t_1$, and cases per gap bin. (b) Cases per question type shared by RTV-Bench and StreamingBench.}

\label{tab:s_stats}
\begin{subtable}[t]{\linewidth}
\centering\setlength{\tabcolsep}{2.6pt}
\caption{Cases and query gaps}\label{tab:s_stats_a}
\begin{tabular*}{\linewidth}{@{\extracolsep{\fill}}l cc cccc@{}}
\toprule
 & & & \multicolumn{4}{c}{\hd{Cases by $t_2-t_1$ (min)}} \\
\cmidrule(l){4-7}
\hd{Source} & \hd{Cases} & \hd{Median gap (s)} & \hd{${\le}3$} & \hd{3--6} & \hd{6--9} & \hd{${>}9$} \\
\midrule
RTV-Bench~\cite{rtvbench2025} & 291 & 120.2 & 179 & 44 & 15 & 53 \\
StreamingBench~\cite{streamingbench2026} & 112 & 335.0 & 19 & 42 & 35 & 16 \\
OVO-Bench~\cite{ovobench2025} & 37 & 105.0 & 22 & 10 & 4 & 1 \\
\midrule
Total & 440 & 181.9 & 220 & 96 & 54 & 70 \\
\bottomrule
\end{tabular*}
\end{subtable}

\vspace{6pt}
\begin{subtable}[t]{\linewidth}
\centering\setlength{\tabcolsep}{1.5pt}
\caption{Question types}\label{tab:s_stats_b}
\begin{tabular*}{\linewidth}{@{\extracolsep{\fill}}l cccc@{}}
\toprule
\hd{Source} & \hd{Perception} & \hd{Understanding} & \hd{Reasoning} & \hd{Counting} \\
\midrule
RTV-Bench~\cite{rtvbench2025} & 82 & 83 & 25 & 101 \\
StreamingBench~\cite{streamingbench2026} & 27 & 56 & 1 & 28 \\
OVO-Bench~\cite{ovobench2025} & 32 & -- & -- & 5 \\
\midrule
Total & 141 & 139 & 26 & 134 \\
\bottomrule
\end{tabular*}
\end{subtable}
\end{table}

\smallskip\noindent\textbf{Statistics.} \Cref{tab:s_stats}(a) summarizes the cases by source. RTV-Bench supplies two thirds of the cases and most of the shortest and longest gaps. \Cref{tab:s_stats}(b) groups the cases into the question types shared by RTV-Bench and StreamingBench.
The RTV-Bench cases alone could drive the overall scores, so \Cref{sec:s_sce_split} reports every result by source. The answer choices keep their order from the source benchmarks, so the prior state $s_1$ is often listed first. \Cref{sec:s_prior} checks whether this affects the results.

\begin{table}[t]
\centering\footnotesize
\setlength{\tabcolsep}{2.0pt}
\caption{\textbf{Accuracy (\%) on \data by source.} $s@t$: question about $s$ asked at $t$. \emph{Cross-time} (\Cone, \Ptwo); \emph{Cross-state} (\Ptwo, \Ctwo). Text only: no video input. RTV: RTV-Bench (291 cases); SB: StreamingBench (112); OVO: OVO-Bench (37).}
\label{tab:s_source}
\begin{tabularx}{\linewidth}{@{}l *{6}{>{\centering\arraybackslash}X}@{}}
\toprule
 & \multicolumn{3}{c}{\hd{Cross-time}} & \multicolumn{3}{c}{\hd{Cross-state}} \\
\cmidrule(lr){2-4}\cmidrule(l){5-7}
\hd{Method} & \hd{RTV} & \hd{SB} & \hd{OVO} & \hd{RTV} & \hd{SB} & \hd{OVO} \\
\midrule
\rowcolor{bbrow}\multicolumn{7}{l}{\textbf{LLaVA-OV-7B}~\cite{llavaonevision2025}} \\
\rf{\quad Text only} & \rf{21.0} & \rf{26.8} & \rf{16.2} & \rf{2.7} & \rf{20.5} & \rf{0.0} \\
\famsep{7}
\plus SimpleStream~\cite{simplestream2026} & 18.2 & 16.1 & 8.1 & 2.7 & 9.8 & 5.4 \\
\famsep{7}
\plus HERMES~\cite{hermes2026} & 22.3 & 20.5 & 16.2 & 4.1 & 16.1 & 2.7 \\
\plus CausalMem~\cite{causalmem2026} & 24.1 & 29.5 & 27.0 & 4.5 & 24.1 & 10.8 \\
\plus CoRDS~\cite{cords2026} & 29.2 & 24.1 & 29.7 & 6.2 & 22.3 & \underline{16.2} \\
\plus InfiniPot-V~\cite{infinipotv2025} & 28.2 & 36.6 & 32.4 & 7.6 & 26.8 & \underline{16.2} \\
\famsep{7}
\plus ReKV~\cite{rekv2025} & \textbf{36.4} & \underline{50.0} & \textbf{40.5} & 7.9 & 33.0 & 13.5 \\
\plus StreamKV~\cite{streamkv2026} & \underline{36.1} & 43.8 & \textbf{40.5} & \underline{8.9} & 31.2 & \underline{16.2} \\
\plus MuKV~\cite{mukv2026} & 25.1 & 33.9 & 35.1 & 3.8 & 18.8 & 2.7 \\
\plus StreamingTOM~\cite{streamingtom2026} & 31.3 & 39.3 & \textbf{40.5} & 6.2 & 25.9 & 10.8 \\
\plus ShallowStream~\cite{shallowstream2026} & 19.9 & 28.6 & 18.9 & 3.8 & 17.0 & 8.1 \\
\rowcolor{oursrow}\plus \textbf{Ours} & 34.7 & \textbf{50.9} & \underline{37.8} & \textbf{11.7} & \underline{47.3} & \textbf{27.0} \\
\rowcolor{oursrow}\plus \textbf{Ours} (INT8) & 34.7 & \textbf{50.9} & \underline{37.8} & \textbf{11.7} & \textbf{48.2} & \textbf{27.0} \\
\midrule
\rowcolor{bbrow}\multicolumn{7}{l}{\textbf{Qwen2.5-VL-7B}~\cite{qwen25vl2025}} \\
\plus SimpleStream~\cite{simplestream2026} & 22.3 & 33.9 & 8.1 & 2.1 & 8.0 & 2.7 \\
\famsep{7}
\plus HERMES~\cite{hermes2026} & 43.0 & 40.2 & \textbf{43.2} & 10.0 & 24.1 & \textbf{27.0} \\
\famsep{7}
\plus ReKV~\cite{rekv2025} & 42.6 & \underline{43.8} & \underline{35.1} & \underline{10.7} & \underline{27.7} & \underline{21.6} \\
\rowcolor{oursrow}\plus \textbf{Ours} & \textbf{44.3} & \textbf{47.3} & 29.7 & \textbf{12.4} & \textbf{30.4} & \underline{21.6} \\
\rowcolor{oursrow}\plus \textbf{Ours} (INT8) & \underline{44.0} & \textbf{47.3} & 29.7 & \textbf{12.4} & \textbf{30.4} & 18.9 \\
\bottomrule
\end{tabularx}
\end{table}

\subsection{Results by Source and Question Type}
\label{sec:s_sce_split}

\smallskip\noindent\textbf{By source.} Since RTV-Bench supplies two thirds of the cases, \Cref{tab:s_source} checks whether it dominates the overall results. 
Ours leads in \emph{Cross-state} on all sources with LLaVA-OV and leads on RTV-Bench and StreamingBench with Qwen2.5-VL.
The trade-off of the main paper also appears on each source. 
Bounded-memory methods and retrieval-based methods score lower on \emph{Cross-time} and \emph{Cross-state}, respectively. 
Hence, the RTV-Bench cases alone do not account for our gain on \data.

\smallskip\noindent\textbf{By question type.} Different question types impose different memory requirements; \Cref{tab:counting} checks whether our gain depends on a particular type. 
With both backbones, ours has the highest \emph{Cross-state} on understanding (U) and counting (C) questions and stays among the best on perception questions (P). 
Retrieval-based methods reach a \emph{Cross-time} comparable to ours on perception and understanding questions but a lower \emph{Cross-state} on understanding questions. 
On counting questions and reasoning cases (R), \emph{Cross-state} stays near the floor for every method.
Our gain is thus not tied to a single question type, although tracking a changing count remains hard for every method.

\begin{table}[t]
\centering\footnotesize
\setlength{\tabcolsep}{2.0pt}
\caption{\textbf{Accuracy (\%) on \data by question type.} P: perception (141 cases), U: understanding (139 cases), R: reasoning (26 cases), and C: counting (134 cases).}
\label{tab:counting}
\begin{tabularx}{\linewidth}{@{}l *{8}{>{\centering\arraybackslash}X}@{}}
\toprule
 & \multicolumn{4}{c}{\hd{Cross-time}} & \multicolumn{4}{c}{\hd{Cross-state}} \\
\cmidrule(lr){2-5}\cmidrule(l){6-9}
\hd{Method} & \hd{P} & \hd{U} & \hd{R} & \hd{C} & \hd{P} & \hd{U} & \hd{R} & \hd{C} \\
\midrule
\rowcolor{bbrow}\multicolumn{9}{l}{\textbf{LLaVA-OV-7B}~\cite{llavaonevision2025}} \\
\rf{\quad Text only} & \rf{22.7} & \rf{28.8} & \rf{11.5} & \rf{16.4} & \rf{5.0} & \rf{12.2} & \rf{7.7} & \rf{3.7} \\
\famsep{9}
\plus SimpleStream~\cite{simplestream2026} & 14.2 & 17.3 & \underline{11.5} & 20.1 & 6.4 & 5.8 & 0.0 & 3.0 \\
\famsep{9}
\plus HERMES~\cite{hermes2026} & 22.0 & 25.2 & 7.7 & 19.4 & 7.1 & 11.5 & 0.0 & 3.7 \\
\plus CausalMem~\cite{causalmem2026} & 29.8 & 28.8 & 3.8 & 22.4 & 9.9 & 16.5 & 0.0 & 5.2 \\
\plus CoRDS~\cite{cords2026} & 34.0 & 28.1 & 7.7 & 25.4 & 15.6 & 12.9 & 0.0 & \underline{6.7} \\
\plus InfiniPot-V~\cite{infinipotv2025} & 37.6 & 38.1 & 3.8 & 20.9 & 19.1 & 17.3 & 0.0 & 5.2 \\
\famsep{9}
\plus ReKV~\cite{rekv2025} & \textbf{45.4} & \textbf{56.1} & \underline{11.5} & 23.9 & 14.2 & \underline{25.2} & \textbf{7.7} & 6.0 \\
\plus StreamKV~\cite{streamkv2026} & 43.3 & 51.1 & 7.7 & \underline{26.1} & 15.6 & \underline{25.2} & \underline{3.8} & \underline{6.7} \\
\plus MuKV~\cite{mukv2026} & 31.9 & 33.8 & 7.7 & 22.4 & 7.1 & 14.4 & \underline{3.8} & 1.5 \\
\plus StreamingTOM~\cite{streamingtom2026} & 40.4 & 44.6 & \underline{11.5} & 20.9 & 14.2 & 18.7 & \underline{3.8} & 3.0 \\
\plus ShallowStream~\cite{shallowstream2026} & 19.9 & 19.4 & 3.8 & \textbf{30.6} & 9.9 & 7.9 & \underline{3.8} & 5.2 \\
\rowcolor{oursrow}\plus \textbf{Ours} & \underline{44.0} & \underline{51.8} & \textbf{15.4} & 25.4 & \underline{19.9} & \textbf{36.7} & \textbf{7.7} & \textbf{11.9} \\
\rowcolor{oursrow}\plus \textbf{Ours} (INT8) & \underline{44.0} & \underline{51.8} & \textbf{15.4} & 25.4 & \textbf{20.6} & \textbf{36.7} & \textbf{7.7} & \textbf{11.9} \\
\midrule
\rowcolor{bbrow}\multicolumn{9}{l}{\textbf{Qwen2.5-VL-7B}~\cite{qwen25vl2025}} \\
\plus SimpleStream~\cite{simplestream2026} & 17.0 & 18.7 & \textbf{11.5} & \textbf{39.6} & 3.5 & 7.2 & 0.0 & 0.7 \\
\famsep{9}
\plus HERMES~\cite{hermes2026} & \textbf{46.1} & 50.4 & \underline{7.7} & \underline{36.6} & 16.3 & \underline{26.6} & \textbf{3.8} & 3.7 \\
\famsep{9}
\plus ReKV~\cite{rekv2025} & \textbf{46.1} & \underline{51.8} & \underline{7.7} & 35.1 & \textbf{19.9} & 23.7 & \textbf{3.8} & \underline{6.0} \\
\rowcolor{oursrow}\plus \textbf{Ours} & \underline{44.0} & \textbf{57.6} & \textbf{11.5} & 35.8 & \underline{17.0} & \textbf{31.7} & \textbf{3.8} & \textbf{6.7} \\
\rowcolor{oursrow}\plus \textbf{Ours} (INT8) & 43.3 & \textbf{57.6} & \textbf{11.5} & 35.8 & 16.3 & \textbf{31.7} & \textbf{3.8} & \textbf{6.7} \\
\bottomrule
\end{tabularx}
\end{table}

\subsection{Answer-Choice Order}
\label{sec:s_prior}
\emph{Always A} ignores the video and picks choice A, the first of the four choices (A/B/C/D), for every question.
Under the source order of the answer choices, used in Table~2 of the main paper, it reaches 49.1\% on \emph{Cross-time}, above every method, but only 5.0\% on \emph{Cross-state} (\Cref{tab:s_shuffle}).
The prior arises because \Cone and \Ptwo share the same answer, most often A, while \Ctwo answers more often occupy later positions.
To avoid this prior, we release \data with shuffled answer choices as its official version.
To shuffle, we randomly permute the answer choices of all 1,320 questions with a fixed seed; \Cone and \Ptwo use the same permutation when they have identical choices, and \Ctwo uses an independent one.
After shuffling, \emph{Always A} drops to 17.7\% on \emph{Cross-time}, whereas that of each method changes by at most 3.5 points.
Ours drops by a few points on \emph{Cross-state} but keeps the highest score, and the order of the methods stays nearly the same, so the source order does not distort the comparison of methods on \data.

\begin{table}[t]
\centering\footnotesize
\setlength{\tabcolsep}{3.0pt}
\caption{\textbf{Accuracy (\%) on \data by answer-choice order.} Orig.: source order; Shuf.: the choices of each question randomly permuted. Always A: picking the first choice for every question.}
\label{tab:s_shuffle}
\begin{tabularx}{\linewidth}{@{}l *{4}{>{\centering\arraybackslash}X}@{}}
\toprule
 & \multicolumn{2}{c}{\hd{Cross-time}} & \multicolumn{2}{c}{\hd{Cross-state}} \\
\cmidrule(lr){2-3}\cmidrule(l){4-5}
\hd{Method} & \hd{Orig.} & \hd{Shuf.} & \hd{Orig.} & \hd{Shuf.} \\
\midrule
\rf{Always A} & \rf{49.1} & \rf{17.7} & \rf{5.0} & \rf{5.2} \\
\famsep{5}
HERMES~\cite{hermes2026} & 21.4 & 23.4 & 7.0 & 8.4 \\
CoRDS~\cite{cords2026} & 28.0 & 24.5 & 11.1 & 10.5 \\
\famsep{5}
ReKV~\cite{rekv2025} & \textbf{40.2} & \textbf{40.2} & \underline{14.8} & \underline{13.9} \\
StreamingTOM~\cite{streamingtom2026} & 34.1 & 35.5 & 11.6 & 10.0 \\
\famsep{5}
\rowcolor{oursrow}\textbf{Ours} & \underline{39.1} & \underline{38.9} & \textbf{22.0} & \textbf{18.6} \\
\bottomrule
\end{tabularx}
\end{table}

\begin{table}[t]
\centering\scriptsize
\setlength{\tabcolsep}{1pt}
\caption{\textbf{Accuracy (\%) on \data by pre-query context length $\ell$.} There are 395, 283, 173, and 106 cases at $\ell=8$, 16, 32, and 64\,s, all evaluated with LLaVA-OV-7B in FP16. \emph{Consistency}: \Ctwo accuracy conditioned on \emph{Cross-time} correct cases.}
\label{tab:s_precontext}
\begin{tabularx}{\linewidth}{@{}l *{12}{>{\centering\arraybackslash}X}@{}}
\toprule
 & \multicolumn{4}{c}{\hd{Cross-time}} & \multicolumn{4}{c}{\hd{Cross-state}} & \multicolumn{4}{c}{\hd{Consistency}} \\
\cmidrule(lr){2-5}\cmidrule(lr){6-9}\cmidrule(l){10-13}
\hd{Method} & \hd{8} & \hd{16} & \hd{32} & \hd{64} & \hd{8} & \hd{16} & \hd{32} & \hd{64} & \hd{8} & \hd{16} & \hd{32} & \hd{64} \\
\midrule
HERMES~\cite{hermes2026} & 17.7 & 14.8 & 15.0 & 15.1 & 5.8 & 4.9 & 5.8 & 2.8 & 22.9 & 23.8 & 15.4 & 6.2 \\
CoRDS~\cite{cords2026} & 19.7 & 18.7 & 16.8 & 13.2 & 5.1 & 6.7 & 4.0 & \underline{3.8} & 21.8 & \underline{32.1} & 17.2 & \underline{21.4} \\
\famsep{13}
ReKV~\cite{rekv2025} & \underline{32.9} & \textbf{30.0} & \underline{25.4} & \textbf{22.6} & \underline{11.1} & \underline{7.1} & \underline{6.4} & \textbf{4.7} & \underline{26.9} & 21.2 & \underline{20.5} & 12.5 \\
StreamingTOM~\cite{streamingtom2026} & 31.1 & 27.2 & \textbf{26.0} & \underline{20.8} & 6.8 & 5.3 & 5.2 & \textbf{4.7} & 17.9 & 15.6 & 17.8 & 13.6 \\
\famsep{13}
\rowcolor{oursrow}\textbf{Ours} & \textbf{34.4} & \underline{29.0} & 22.5 & 19.8 & \textbf{15.4} & \textbf{12.4} & \textbf{7.5} & \textbf{4.7} & \textbf{40.4} & \textbf{34.1} & \textbf{28.2} & \textbf{23.8} \\
\bottomrule
\end{tabularx}
\end{table}

\subsection{Pre-Query Context Length}
\label{sec:s_precontext}
\Cref{tab:s_precontext} lists the exact values plotted in Figure~6 of the main paper.
Each case needs at least $\ell$\,s of source video before $t_1$ in addition to $t_1$ falling within the first quarter of the clip, so fewer cases remain at longer lengths.
\emph{Cross-time} requires recalling $s_1$ after it changes, and the bounded-memory baselines score lower on it than the retrieval-based ones, which indicates that they lose the prior state more often.
\emph{Cross-state} replaces \Cone with the current-state question \Ctwo, and on it this lead shrinks, showing that recalling the prior state often does not come with tracking the current one.
Both weaknesses appear at every length, so the findings on \data do not depend on $\ell$.
These results highlight the need for a standalone benchmark with annotated state changes.

\subsection{Prefix Sensitivity}
\label{sec:s_robust}
Since \Ptwo refers to the prior state only through its prefix, \Cref{tab:s_prefix} checks whether the results depend on its prefix. We replace the default ``Near the beginning of this clip'' (P0) with ``At the start of this clip'' (P1), ``In the first seconds of this clip'' (P2), ``In the opening of this clip'' (P3), and ``Close to the beginning of this clip'' (P4).
The ranking is the same under every prefix: ours has the highest \emph{Cross-state} and \emph{Consistency}, and ReKV the highest \emph{Cross-time} with ours second.
Even our lowest \emph{Cross-state} (18.6) stays above the highest \emph{Cross-state} of any baseline under any prefix (15.7, ReKV with P1 and P3). These results indicate that our gain does not depend on the exact prefix.

\begin{table}[t]
\centering\scriptsize
\setlength{\tabcolsep}{1pt}
\caption{\textbf{Accuracy (\%) on \data by prefix.} P1--P4: the alternative prefixes for \Ptwo. \emph{Consistency}: \Ctwo accuracy conditioned on \emph{Cross-time} correct cases.}
\label{tab:s_prefix}
\begin{tabularx}{\linewidth}{@{}l *{12}{>{\centering\arraybackslash}X}@{}}
\toprule
 & \multicolumn{4}{c}{\hd{Cross-time}} & \multicolumn{4}{c}{\hd{Cross-state}} & \multicolumn{4}{c}{\hd{Consistency}} \\
\cmidrule(lr){2-5}\cmidrule(lr){6-9}\cmidrule(l){10-13}
\hd{Method} & \hd{P1} & \hd{P2} & \hd{P3} & \hd{P4} & \hd{P1} & \hd{P2} & \hd{P3} & \hd{P4} & \hd{P1} & \hd{P2} & \hd{P3} & \hd{P4} \\
\midrule
HERMES~\cite{hermes2026} & 20.9 & 21.6 & 21.6 & 21.1 & 8.2 & 7.5 & 8.9 & 6.8 & 31.5 & 28.4 & 32.6 & 25.8 \\
CoRDS~\cite{cords2026} & 26.4 & 27.0 & 27.0 & 27.0 & 11.1 & 11.4 & 11.8 & 10.0 & \underline{38.8} & \underline{37.8} & \underline{40.3} & \underline{35.3} \\
\famsep{13}
ReKV~\cite{rekv2025} & \textbf{40.0} & \textbf{39.1} & \textbf{40.2} & \textbf{39.8} & \underline{15.7} & \underline{15.5} & \underline{15.7} & \underline{14.8} & 32.4 & 32.6 & 32.8 & 30.9 \\
StreamingTOM~\cite{streamingtom2026} & 34.5 & 34.8 & 33.6 & 34.3 & 10.9 & 12.3 & 10.9 & 12.0 & 27.6 & 28.1 & 27.0 & 27.8 \\
\famsep{13}
\rowcolor{oursrow}\textbf{Ours} & \underline{37.5} & \underline{36.1} & \underline{38.0} & \underline{38.4} & \textbf{19.8} & \textbf{18.6} & \textbf{20.2} & \textbf{20.9} & \textbf{48.5} & \textbf{46.5} & \textbf{47.9} & \textbf{48.5} \\
\bottomrule
\end{tabularx}
\end{table}

\begin{table}[t]
\centering\footnotesize
\setlength{\tabcolsep}{2.0pt}
\caption{\textbf{Accuracy (\%) on \data with smaller backbones.} Settings follow the 7B backbones, except that the vision encoder of Qwen2.5-VL-3B runs in FP32 for every method.}
\label{tab:s_ov05b}
\begin{tabular}{@{}l c cc cc c@{}}
\toprule
\multirow{2}{*}[-1.5pt]{\hd{Method}} & \hd{$@t_1$} & \multicolumn{2}{c}{\hd{$@t_2$}} & \multicolumn{2}{c}{\hd{Cross-}} & \multirow{2}{*}[-1.5pt]{\hd{Consistency}} \\[-1pt]
\cmidrule(lr){2-2}\cmidrule(lr){3-4}\cmidrule(lr){5-6}
 & \hd{$s_1$} & \hd{$s_1$} & \hd{$s_2$} & \hd{time} & \hd{state} & \\
\midrule
\rowcolor{bbrow}\multicolumn{7}{l}{\textbf{LLaVA-OV-0.5B}} \\
SimpleStream~\cite{simplestream2026} & 52.5 & 29.8 & 43.0 & 26.6 & 3.2 & 12.0  \\
\famsep{7}
HERMES~\cite{hermes2026} & 50.9 & 33.6 & 38.4 & 28.9 & 4.3 & 12.6  \\
CausalMem~\cite{causalmem2026} & 49.8 & 33.6 & 37.0 & 27.5 & 4.1 & 13.2  \\
CoRDS~\cite{cords2026} & 52.5 & 35.5 & 34.8 & 27.0 & 3.6 & 10.1  \\
InfiniPot-V~\cite{infinipotv2025} & 49.8 & 35.7 & 36.4 & 29.1 & 5.7 & 13.3  \\
\famsep{7}
ReKV~\cite{rekv2025} & 50.0 & 39.1 & 32.3 & \underline{32.5} & 5.5 & 14.0  \\
StreamKV~\cite{streamkv2026} & 49.1 & 34.8 & 33.4 & 28.9 & 4.3 & 13.4  \\
MuKV~\cite{mukv2026} & 49.8 & 39.8 & 26.8 & 30.2 & 5.9 & \underline{16.5}  \\
StreamingTOM~\cite{streamingtom2026} & 51.8 & 40.7 & 31.8 & \textbf{33.0} & \underline{6.1} & 15.2  \\
ShallowStream~\cite{shallowstream2026} & 52.5 & 32.7 & 42.5 & 29.3 & 5.0 & 16.3  \\
\famsep{7}
\rowcolor{oursrow}\ours~\textbf{(Ours)} & 50.0 & 35.5 & 40.5 & 28.9 & \textbf{7.0} & \textbf{18.1}  \\
\rowcolor{bbrow}\multicolumn{7}{l}{\textbf{Qwen2.5-VL-3B}} \\
SimpleStream~\cite{simplestream2026} & 47.7 & 28.9 & 47.7 & 20.7 & 6.8 & 22.0 \\
\famsep{7}
HERMES~\cite{hermes2026} & 48.6 & 41.6 & 40.5 & \underline{33.4} & \textbf{13.2} & \underline{32.0} \\
\famsep{7}
ReKV~\cite{rekv2025} & 50.2 & 40.5 & 40.9 & \underline{33.4} & \textbf{13.2} & \textbf{33.3} \\
\rowcolor{oursrow}\ours~\textbf{(Ours)} & 50.2 & 42.0 & 41.8 & \textbf{35.2} & \underline{12.5} & 31.0 \\
\bottomrule
\end{tabular}
\end{table}

\subsection{Smaller Backbones}
\label{sec:s_ov05b}
\Cref{tab:s_ov05b} shows the \data evaluation with LLaVA-OV-0.5B and Qwen2.5-VL-3B. 
Only the gate layers move to the same relative depth (layers 9--15 of 24 and 13--23 of 36, respectively), and $\tau$ remains unchanged from the 7B setting. 
With both smaller backbones, paired accuracy drops for every method, and the gaps between methods shrink.
With LLaVA-OV-0.5B, ours keeps the highest \emph{Cross-state}, but its lead shrinks the most and its \emph{Cross-time} falls below that of most retrieval-based methods.
With Qwen2.5-VL-3B, ours has the highest \emph{Cross-time} and stays within one point of the best \emph{Cross-state}.
On this backbone, the gate looks back on 98.0\% of the questions and rarely takes the recent-only route.
Looking back on nearly every question lowers \Ctwo accuracy, as \Cref{fig:s_tau} shows for LLaVA-OV, and with it \emph{Cross-state}, showing that routing matters.

\section{Method Details}
\label{sec:s_method}
\Cref{alg:s_write} shows how a frame that leaves the recent window is stored, and \Cref{alg:s_group,alg:s_factor,alg:s_summ,alg:s_share} detail its steps. A group first tries to share the statistics of an earlier group and is factorized as a new parent only when no parent satisfies the sharing criterion. All steps run for each layer; groups are shared across KV heads, and statistics are kept per KV head. \Cref{alg:s_read} gives details of the look-back gate, and \Cref{tab:s_impl} lists the settings for LLaVA-OV and Qwen2.5-VL.

\begin{algorithm}[t]
\caption{\ours Memory Write}
\label{alg:s_write}
\midsize
\begin{algorithmic}
\Require \AlgLines{frame $f$ leaving the recent window, at one layer \\ keys and values $\{\mathbf{x}_i\}_{i=1}^{N}$; shared basis $\mathbf{U}$}
\Ensure record of $f$
\Stage{1. Grouping:}
  \State $\mathcal{G}\gets\Mod{Grouping}(f)$ \Comment{\Cref{alg:s_group}}
\EndStage
\For{$g\in\mathcal{G}$}
  \Stage{2. Cross-Frame Sharing or Factorizing:}
    \State $p\gets\Mod{Cross-Frame Sharing}(g)$ \Comment{\Cref{alg:s_share}}
    \If{$p=\varnothing$} \Comment{new parent, \Cref{alg:s_factor}}
      \State $(\boldsymbol{\mu}_g,\boldsymbol{\sigma}_g,\boldsymbol{\Phi}_g,\{\mathbf{c}_i\})\gets\Mod{Factorizing}(g)$
    \Else
      \State $(\boldsymbol{\mu}_g,\boldsymbol{\sigma}_g,\boldsymbol{\Phi}_g)\gets(\boldsymbol{\mu}_p,\boldsymbol{\sigma}_p,\boldsymbol{\Phi}_p)$
      \State $\{\mathbf{c}_i\},\boldsymbol{\phi}_g,\{\zeta_i\}\gets$ outputs of \Cref{alg:s_share}
    \EndIf
  \EndStage
  \Stage{3. Summarizing:}
    \State $\{(\mathbf{r}_k,w_k,\eta_k)\}\gets\Mod{Summarizing}(g)$ \Comment{\Cref{alg:s_summ}}
  \EndStage
\EndFor
\Stage{4. Storing:}
  \State keep $\{(\mathbf{r}_k,w_k,\eta_k)\}$ and their spatial positions
  \State keep $(\boldsymbol{\mu}_g,\boldsymbol{\sigma}_g,\boldsymbol{\Phi}_g)$ of each new parent
  \State keep $\boldsymbol{\phi}_g$ of each shared group
  \State keep the retrieval key of $f$ \Comment{\Cref{alg:s_read}}
\EndStage
\end{algorithmic}
\end{algorithm}

\begin{algorithm}[t]
\caption{Grouping}
\label{alg:s_group}
\midsize
\begin{algorithmic}
\Require \AlgLines{keys and values of the $N$ tokens of $f$ \\ split threshold $\lambda$; at most four rounds}
\Ensure groups $\mathcal{G}$, shared by all KV heads
\Stage{1. Joint KV feature:}
  \State $\bar{\mathbf{k}}_i\gets$ keys of token $i$ stacked over KV heads
  \State $\bar{\mathbf{v}}_i\gets$ values of token $i$ stacked over KV heads
  \State center $\bar{\mathbf{k}}$, $\bar{\mathbf{v}}$ over tokens 
  \State scale each to unit root mean square (RMS)
  \State $\mathbf{z}_i\gets[\bar{\mathbf{k}}_i,\bar{\mathbf{v}}_i]$;\; $\mathcal{G}\gets\{\{1,\ldots,N\}\}$
\EndStage
\Stage{2. Bisecting $k$-means (BKM):}
\For{round $=1,\ldots,4$}
    \For{$g\in\mathcal{G}$}
      \State split $g$ into $g_a,g_b$ by $2$-means on $\{\mathbf{z}_i\}_{i\in g}$
      \State $\Delta J\gets J(g)-J(g_a)-J(g_b)$ \Comment{$J$: squared error}
      \If{$\Delta J/\dim(\mathbf{z})>\lambda$} replace $g$ by $g_a,g_b$ \EndIf
    \EndFor
    \If{no split was accepted} \textbf{break} \EndIf
  \EndFor
  \State reassign all $\mathbf{z}_i$ by 4 Lloyd steps from the group means
\EndStage
\State \Return $\mathcal{G}$ \Comment{at most 16 groups}
\end{algorithmic}
\end{algorithm}

\subsection{Grouping}
\label{sec:s_group}
\Cref{alg:s_group} gives the grouping procedure. Each 2-means split starts from the token farthest from the group mean and the token farthest from that one, and runs four Lloyd iterations; a group with a single token is not split. A split is kept only if both halves are non-empty and the squared error per feature dimension drops by more than $\lambda$, so a frame with uniform content stays in few groups. After the split rounds, four Lloyd iterations of $k$-means over all tokens, starting from the group means, refine the assignments, with a repair step for groups that become empty. With at most four rounds, a frame has at most 16 groups per layer, and the split threshold $\lambda$ decides how many of them are formed.

\begin{algorithm}[t]
\caption{Factorizing}
\label{alg:s_factor}
\midsize
\begin{algorithmic}
\Require \AlgLines{tokens $\{\mathbf{x}_i\}_{i\in g}$ of a new parent $g$ at one KV head \\ basis dimensions $D_U$, $D_\Phi$}
\Ensure center $\boldsymbol{\mu}_g$, scales $\boldsymbol{\sigma}_g$, basis $\boldsymbol{\Phi}_g$, coefficients $\{\mathbf{c}_i\}$
\Stage{1. Shared basis (once per video, at frame $W$):}
  \State $\mathbf{E}\gets$ stacked $\mathbf{e}_i$ of all groups in the first $W$ frames
  \State $\mathbf{U}\gets\Mod{PivChol}(\mathbf{E}^\top\mathbf{E},D_U)$ \Comment{\Cref{eq:s_pchol}}
  \State $\mathbf{U}\gets\mathrm{CholeskyQR2}(\mathbf{U})$
\EndStage
\Stage{2. Normalization:}
  \State $\boldsymbol{\mu}_g,\boldsymbol{\sigma}_g\gets$ center and key/value scales of $g$
  \State $\mathbf{e}_i\gets(\mathbf{x}_i-\boldsymbol{\mu}_g)\oslash\boldsymbol{\sigma}_g$
\EndStage
\Stage{3. Group basis:}
  \State $\mathbf{y}_i\gets\mathbf{U}^\top\mathbf{e}_i$;\; $\mathbf{S}_g\gets\frac{1}{N_g}\sum_{i\in g}\mathbf{y}_i\mathbf{y}_i^\top$
  \State $\boldsymbol{\Phi}_g\gets\Mod{PivChol}(\mathbf{S}_g,\min(D_\Phi,N_g{-}1))$ \Comment{early stop}
  \State $\boldsymbol{\Phi}_g\gets\mathrm{CholeskyQR2}(\boldsymbol{\Phi}_g)$, stored in FP16
\EndStage
\Stage{4. Coefficients:}
  \State $\mathbf{c}_i\gets\arg\min_{\mathbf{c}}\lVert\mathbf{y}_i-\boldsymbol{\Phi}_g\mathbf{c}\rVert$ \Comment{FP16 $\boldsymbol{\Phi}_g$}
\EndStage
\State \Return $\boldsymbol{\mu}_g,\boldsymbol{\sigma}_g,\boldsymbol{\Phi}_g,\{\mathbf{c}_i\}$
\end{algorithmic}
\end{algorithm}

\subsection{Factorizing}
\label{sec:s_factor}
\smallskip\noindent\textbf{Pivoted Cholesky factorization.} Factorizing fits two bases with the same routine, \Mod{PivChol} in \Cref{alg:s_factor}: the shared basis $\mathbf{U}$ of each layer and KV head (stage 1) and the group basis $\boldsymbol{\Phi}_g$ of each new parent (stage 3), which later groups can reuse through cross-frame sharing (\Cref{alg:s_share}).
\Mod{PivChol} builds a low-rank factor of a second-moment matrix one column at a time.
At each step, it picks the coordinate $j$ with the largest remaining diagonal entry, takes the matching column of the remaining matrix as a new direction, and removes that direction's contribution before the next step:
\begin{equation}
  \boldsymbol{\ell}=[\tilde{\mathbf{S}}]_{:,j}\big/\sqrt{[\tilde{\mathbf{S}}]_{jj}},\qquad \tilde{\mathbf{S}}\leftarrow\tilde{\mathbf{S}}-\boldsymbol{\ell}\boldsymbol{\ell}^\top,
  \label{eq:s_pchol}
\end{equation}
where $\tilde{\mathbf{S}}$ starts from the second-moment matrix.
Unlike an eigendecomposition, it reads only the diagonal and one column of the matrix per step.
Since the chosen directions are not orthogonal, CholeskyQR2 orthonormalizes them.

\smallskip\noindent\textbf{Shared and group bases.} The shared basis $\mathbf{U}$ is fitted once per video on the normalized residuals of all groups in the first $W$ frames and always keeps all $D_U$ directions.
The group basis $\boldsymbol{\Phi}_g$ is fitted on the projected residuals $\mathbf{y}_i$ of its group, and its second-moment matrix is never formed explicitly, since each step needs only the diagonal and one column.
It keeps at most $\min(D_\Phi,N_g{-}1)$ directions, because the centered residuals of $N_g$ tokens span at most $N_g{-}1$ dimensions, and stops early once the largest remaining diagonal entry falls below $10^{-6}$ of the initial trace.
$\boldsymbol{\Phi}_g$ is stored in FP16, and the coefficients $\mathbf{c}_i$ are solved by least squares against the stored basis, ensuring that rounding of the basis does not bias them.

\subsection{Summarizing}
\label{sec:s_summ}
\begin{algorithm}[t]
\caption{Summarizing}
\label{alg:s_summ}
\midsize
\begin{algorithmic}
\Require \AlgLines{coefficients $\{\mathbf{c}_i\}_{i\in g}$ and positions of the $N_g$ tokens \\ residual coefficients $\{\zeta_i\}$ ($0$ for a new parent) \\ at most $M$ summary tokens}
\Ensure summary tokens $\mathbf{r}_k$, weights $w_k$, coefficients $\eta_k$
\If{$N_g\le M$} \Comment{keep every token}
  \State $\mathbf{r}_k\gets\mathbf{c}_k$;\; $w_k\gets1$;\; $\eta_k\gets\zeta_k$
\Else
  \Stage{1. Token selection:}
    \State $\mathcal{C}_1,\ldots,\mathcal{C}_M\gets$ $k$-means on $\{\mathbf{c}_i\}$
    \State $\mathbf{r}_k\gets$ the $\mathbf{c}_i$ in $\mathcal{C}_k$ nearest to its center
    \State $w_k\gets|\mathcal{C}_k|$; keep the spatial position of $\mathbf{r}_k$
  \EndStage
  \Stage{2. Moment correction:}
    \State $\bar{\mathbf{c}},\boldsymbol{\Sigma}\gets$ mean and covariance of $\{\mathbf{c}_i\}$
    \State $\bar{\mathbf{r}},\boldsymbol{\Sigma}_r\gets$ weighted mean and covariance of $\{\mathbf{r}_k\}$
    \State $\mathbf{L}_c\gets$ Cholesky factor: $\boldsymbol{\Sigma}=\mathbf{L}_c\mathbf{L}_c^\top$
    \State $\mathbf{L}_r\gets$ Cholesky factor: $\boldsymbol{\Sigma}_r=\mathbf{L}_r\mathbf{L}_r^\top$
    \State $\mathbf{r}_k\gets\bar{\mathbf{c}}+\mathbf{L}_c\mathbf{L}_r^{-1}(\mathbf{r}_k-\bar{\mathbf{r}})$ \Comment{\Cref{eq:s_moment}}
  \EndStage
  \Stage{3. Residual coefficients (shared groups):}
    \State fit $\{\eta_k\}$ to the mean and variance of $\{\zeta_i\}$
    \State and to the covariance of $\{\zeta_i\}$ with $\{\mathbf{c}_i\}$
  \EndStage
\EndIf
\State \Return $\{(\mathbf{r}_k,w_k,\eta_k)\}$
\end{algorithmic}
\end{algorithm}

\smallskip\noindent\textbf{Token selection.} We keep every token of a group with at most $M$ tokens.
For a larger group, stage 1 of \Cref{alg:s_summ} clusters the token coefficients with $k$-means and keeps, from each cluster, the actual token nearest to its center, weighted by the cluster size.
Each summary token therefore carries the spatial position of a real token, which the readout uses for RoPE, and the weights add up to the number of tokens in the group.

\smallskip\noindent\textbf{Moment correction.} Since $k$-means keeps only the spread between clusters, the kept tokens are typically less spread out than the group.
Stage 2 therefore moves them by an affine map so that their weighted mean and covariance match those of all token coefficients:
\begin{equation}
  \mathbf{r}_k\leftarrow\bar{\mathbf{c}}+\mathbf{L}_c\mathbf{L}_r^{-1}(\mathbf{r}_k-\bar{\mathbf{r}}),
  \label{eq:s_moment}
\end{equation}
where $\bar{\mathbf{c}}$ and $\mathbf{L}_c$ come from the token coefficients and $\bar{\mathbf{r}}$ and $\mathbf{L}_r$ from the kept tokens under their weights. $\mathbf{L}_r^{-1}$ removes the spread of the kept tokens, and $\mathbf{L}_c$ gives them the spread of the group.
The map changes only the coefficients, so each summary token keeps its position.
$\mathbf{L}_r$ is invertible only if the kept tokens span all $D_\Phi$ directions, which requires $M>D_\Phi$ (\ie, $M{=}8>D_\Phi{=}4$ in our setting); otherwise, the group places $2D_\Phi$ fixed points around its mean instead.
Stage 3 fits the coefficients of the extra residual direction, which keeps the part of a shared group's residuals that the parent basis misses, to limit loss from sharing statistics; it is described with cross-frame sharing in \Cref{sec:s_share}.

\subsection{Cross-Frame Sharing}
\label{sec:s_share}
\begin{algorithm}[t]
\caption{Cross-Frame Sharing}
\label{alg:s_share}
\midsize
\begin{algorithmic}
\Require \AlgLines{group $g$ of the current frame \\ pool $\mathcal{P}$: groups of the previous frame and \\ the 256 most recently stored groups}
\Ensure \AlgLines{parent $p$, coefficients $\{\mathbf{c}_i\}$, direction $\boldsymbol{\phi}_g$, \\ residual coefficients $\{\zeta_i\}$; or $p{=}\varnothing$ (new parent)}
\Stage{1. Candidate selection:}
   \State $p\gets\arg\min_{p'\in\mathcal{P}}\lVert\boldsymbol{\mu}_g-\boldsymbol{\mu}_{p'}\rVert$
  \If{$g\ne\arg\min_{g'\in\mathcal{G}}\lVert\boldsymbol{\mu}_{g'}-\boldsymbol{\mu}_p\rVert$} \Return $\varnothing$ \EndIf
\EndStage
\Stage{2. Sharing decision:}
  \State $\mathbf{e}_i^p\gets(\mathbf{x}_i-\boldsymbol{\mu}_p)\oslash\boldsymbol{\sigma}_p$
  \State $\hat{\mathbf{x}}_i\gets\boldsymbol{\mu}_p+\boldsymbol{\sigma}_p\odot\mathbf{U}\boldsymbol{\Phi}_p\boldsymbol{\Phi}_p^\top\mathbf{U}^\top\mathbf{e}_i^p$ \Comment{reconstruction}
  \State $\epsilon_i\gets$ larger over keys and values of
  \State \quad $\lVert\mathbf{x}_i-\hat{\mathbf{x}}_i\rVert/(\sqrt{D}\,\lVert\boldsymbol{\sigma}_p\rVert)$ \Comment{all KV heads}
  \If{$\frac{1}{N_g}\sum_{i\in g}\epsilon_i^2>1$} \Return $\varnothing$ \EndIf
\EndStage
\Stage{3. Extra residual correction:}
  \State $\mathbf{y}_i^{p}\gets\mathbf{U}^\top\mathbf{e}_i^{p}$;\; $\mathbf{c}_i\gets\boldsymbol{\Phi}_p^\top\mathbf{y}_i^{p}$
  \State $\mathbf{y}_i^{\perp}\gets(\mathbf{I}-\boldsymbol{\Phi}_p\boldsymbol{\Phi}_p^\top)\mathbf{y}_i^{p}$
  \State $\boldsymbol{\phi}_g\gets$ top eigenvector of $\sum_{i\in g}\mathbf{y}_i^{\perp}\mathbf{y}_i^{\perp\top}$
  \State $\zeta_i\gets\boldsymbol{\phi}_g^\top\mathbf{y}_i^{\perp}$
\EndStage
\State \Return $p,\{\mathbf{c}_i\},\boldsymbol{\phi}_g,\{\zeta_i\}$
\end{algorithmic}
\end{algorithm}

\smallskip\noindent\textbf{Candidate selection.} \Cref{alg:s_share} first selects a candidate parent among recent groups, namely those of the previous frame and the 256 most recently stored ones.
The candidate is the group whose center is closest to the center of the current group, provided that the current group is in turn the closest to it among the groups of the current frame (stage 1).
The distance between two centers uses only their key halves, concatenated over KV heads.

\smallskip\noindent\textbf{Sharing decision.} Stage 2 normalizes each token of the current group with the center $\boldsymbol{\mu}_p$ and scales $\boldsymbol{\sigma}_p$ of the parent, giving its residual $\mathbf{e}_i^p$, then projects this residual onto the parent basis and undoes the normalization, giving the reconstruction $\hat{\mathbf{x}}_i$.
The error $\epsilon_i$ is the distance between $\mathbf{x}_i$ and $\hat{\mathbf{x}}_i$ over all KV heads, divided by $\sqrt{D}\,\lVert\boldsymbol{\sigma}_p\rVert$, the RMS distance of the parent's tokens from $\boldsymbol{\mu}_p$, since $\boldsymbol{\sigma}_p$ holds one key scale and one value scale per KV head.
It is computed for keys and values separately, and the larger one is kept so that a group shares only if both its keys and values fit the parent.
The group shares the statistics of the parent if the mean of $\epsilon_i^2$ is at most one.
Otherwise, it is factorized as a new parent through Factorizing (\Cref{alg:s_factor}).

\smallskip\noindent\textbf{Extra residual correction.} A shared group uses the parent basis, leaving the part of its residuals outside that basis unrepresented.
Stage 3 keeps the main direction of this part, the top eigenvector $\boldsymbol{\phi}_g$ of the out-of-basis residuals, and one coefficient $\zeta_i$ per token along it.
A group of at most $M$ tokens stores these coefficients unchanged, and a summarized group sets one coefficient $\eta_k$ per summary token so that the pairs $(\mathbf{r}_k,\eta_k)$ have the weighted mean and covariance of $(\mathbf{c}_i,\zeta_i)$ (stage 3 of \Cref{alg:s_summ}), in the same way as the moment correction.
A new parent stores no extra direction.

\begin{algorithm}[t]
\caption{Look-Back Gate}
\label{alg:s_read}
\midsize
\begin{algorithmic}
\Require \AlgLines{question $q$ with its answer choices \\ $\mathcal{R}$: the last $\lfloor W/2\rfloor$ recent frames \\ layers $\mathcal{L}$; threshold $\tau$}
\Ensure answer to $q$, or a request to look back
\Stage{1. Two copies:}
  \State $q^{\mathrm{rec}}\gets$ copy of $q$ placed after $\mathcal{R}$
  \State $q^{\mathrm{txt}}\gets$ copy of $q$ masked from $\mathcal{R}$ (text only)
  \State run $q^{\mathrm{rec}}$ and $q^{\mathrm{txt}}$ in one forward pass
  \State $\mathbf{h}^{\mathrm{rec}}_l,\mathbf{h}^{\mathrm{txt}}_l\gets$ last-token states of $q^{\mathrm{rec}},q^{\mathrm{txt}}$, $l\in\mathcal{L}$
\EndStage
\Stage{2. Routing:}
  \State $\delta\gets\frac{1}{|\mathcal{L}|}\sum_{l\in\mathcal{L}}\bigl(1-\cos(\mathbf{h}^{\mathrm{rec}}_l,\mathbf{h}^{\mathrm{txt}}_l)\bigr)$
  \If{$\delta>\tau$} \Comment{recent-only route}
    \State \Return answer decoded from $q^{\mathrm{rec}}$
  \Else \Comment{look-back route}
    \State \Return look back (retrieval and readout)
  \EndIf
\EndStage
\end{algorithmic}
\end{algorithm}

\begin{figure}[t]
\centering
\includegraphics[width=0.9\linewidth]{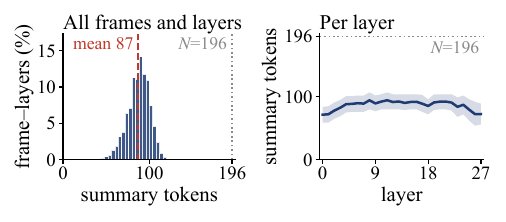}
\caption{\textbf{Summary tokens per stored frame.} Distribution over all frames and layers (left) and mean with 10--90\% range per layer (right) on \data (LLaVA-OV). Dotted lines: the $N{=}196$ visual tokens of a frame.}
\label{fig:s_ntok}
\end{figure}

\subsection{Look-Back Gate, Retrieval, and Readout}
\label{sec:s_read}
\smallskip\noindent\textbf{Look-back gate.} \Cref{alg:s_read} decides whether a question needs past frames. Of the two copies of the question, $q^{\mathrm{rec}}$ follows the recent frames $\mathcal{R}$, and $q^{\mathrm{txt}}$ is masked from them. A large $\delta$ means that the recent frames change how the model reads the question, so the answer is decoded from $q^{\mathrm{rec}}$; a small $\delta$ means that they add little, so the gate looks back. We use the intermediate layers $\mathcal{L}$ (\Cref{tab:s_impl}) because visual information is integrated into the text tokens by the middle layers of multimodal LLMs~\cite{kaduri2025vision,zhang2025crossmodal}.

\smallskip\noindent\textbf{Retrieval and readout.} When the gate looks back, we retrieve past frames as in ReKV~\cite{rekv2025}, using the mean key of each stored frame as its retrieval key (\Cref{tab:s_impl}).
Each retrieved frame is read as its summary tokens, decoded as $\hat{\mathbf{x}}_k=\boldsymbol{\mu}_p+\boldsymbol{\sigma}_p\odot\mathbf{U}(\boldsymbol{\Phi}_p\mathbf{r}_k+\boldsymbol{\phi}_g\eta_k)$ ($p{=}g$ and $\eta_k{=}0$ for a new parent), with RoPE applied to their keys at the frame's position in the context and the stored spatial positions, and $\log w_k$ added to their attention logits.
On \data, a retrieved frame is read as about 90 summary tokens per layer instead of 196 visual tokens (\Cref{fig:s_ntok}).

\section{Implementation Details}
\label{sec:s_impl}

\begin{table}[t]
\centering
\footnotesize
\setlength{\tabcolsep}{3.6pt}
\caption{\textbf{Implementation settings.} Unit: frames encoded as one visual feature. BKM: bisecting $k$-means per frame and layer with at most four split rounds.}
\label{tab:s_impl}
\begin{tabular}{l cc}
\toprule
\hd{Setting} & \hd{LLaVA-OV} & \hd{Qwen2.5-VL} \\
\midrule
\multicolumn{3}{l}{\emph{Backbone and stream processing}} \\
\quad Sampling rate & 1\,fps & 1\,fps \\
\quad Pixels per frame & 384$\times$384 & ${\le}$147,456 \\
\quad Unit & 1 frame & 2 frames \\
\quad Recent window $W$ & 10 frames & 5 units \\
\quad GPU & RTX A6000 & RTX A6000 \\
\midrule
\multicolumn{3}{l}{\emph{Memory}} \\
\quad Grouping & BKM & BKM \\
\quad Split threshold $\lambda$ & 2.5 & 2.5 \\
\quad Shared basis width $D_U$ & 48 & 48 \\
\quad Group basis width $D_\Phi$ & 4 & 4 \\
\quad Summary tokens per group $M$ & ${\le}$8 & ${\le}$8 \\
\quad Precision & FP16 (INT8) & FP16 (INT8) \\
\midrule
\multicolumn{3}{l}{\emph{Look-back gate}} \\
\quad Gate context & last 5 frames & last 2 units \\
\quad Layers $\mathcal{L}$ & 10--18 & 10--18 \\
\quad Threshold $\tau$ & 0.015 & 0.010 \\
\midrule
\multicolumn{3}{l}{\emph{Retrieval and readout}} \\
\quad Retrieved past frames & ${\le}$54 frames & ${\le}$27 units \\
\quad Recent-only route & last 5 frames & last 2 units \\
\quad Retrieval key (mean per unit) & pre-RoPE & post-RoPE \\
\bottomrule
\end{tabular}
\end{table}

\Cref{tab:s_impl} lists our settings on LLaVA-OV and Qwen2.5-VL, whose budgets are counted in units of two frames. The baselines follow the configuration of each paper and its released code with the default budget. All methods see the same 1\,fps frames and questions and answer greedily with at most 8 new tokens. The settings below are for LLaVA-OV unless noted, and the configuration of every method will be released with our code.

\smallskip\noindent\textbf{Recent-frame baseline.} SimpleStream answers from the most recent frames only and keeps no memory of earlier ones. It keeps the last 5 frames with LLaVA-OV and the last 4 frames (2 units) with Qwen2.5-VL.

\smallskip\noindent\textbf{Bounded-memory baselines.} These methods keep a cache of fixed size by selecting, merging, or summarizing tokens as the stream grows. 
HERMES keeps a 6K-token KV cache filled in chunks of 16 frames on both backbones, and CausalMem keeps a 12K-token memory. 
CoRDS uses the setting of its paper (the bottom 25\% of layers, blocks of 32 frames compressed to 24), and InfiniPot-V compresses a 32-frame cache to 24 frames with a query ratio of 0.125.

\smallskip\noindent\textbf{Retrieval-based baselines.} These methods keep a growing archive of past frames and retrieve part of it for each question. ReKV, MuKV, and StreamKV retrieve 64 past frames or segments; ReKV and MuKV keep a 15K-token local window, and StreamKV keeps 40\% of the tokens of each segment. 
StreamingTOM uses its released 4-bit memory, and ShallowStream its released configuration without long-term clustering, which keeps the input embedding of every past frame. 
With Qwen2.5-VL, ReKV retrieves 32 units (64 frames) with the same retrieval key as ours, so both read at most 32 units per question.

\begin{figure}[t]
\centering
\includegraphics[width=\linewidth]{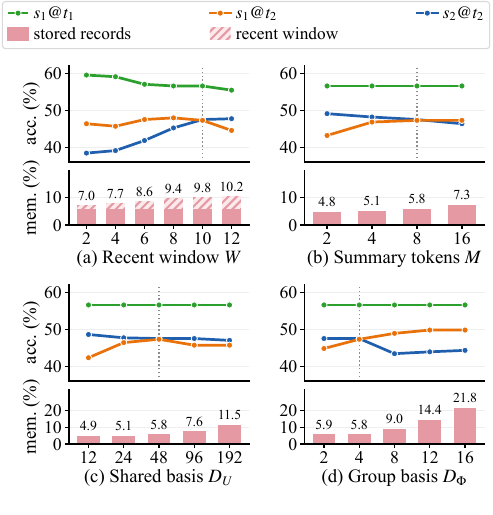}
\caption{\textbf{Hyperparameter sensitivity.} Accuracy (\%) on \data by question and memory (\% of raw KV, including retrieval keys) of ours for each hyperparameter. Hatched bar in (a): the recent window, which the reported memory excludes. The gate uses the last $W/2$ frames. Dotted lines: our setting.}
\label{fig:s_sens}
\end{figure}

\section{Further Analysis}
\label{sec:s_analysis}

\subsection{Hyperparameter Sensitivity}
\label{sec:s_sens}
\smallskip\noindent\textbf{Recent window $W$ and summary tokens $M$.} \Cref{fig:s_sens}(a, b) varies $W$ and $M$ around our setting.
As $W$ grows, \Ctwo rises and saturates at our setting.
With few recent frames, the question representation changes little, leading the gate to look back on almost every question (99.8\% at $W{=}2$), which reduces current-state accuracy; beyond our setting, the gate looks back on only half of the questions, reducing accuracy on \Ptwo.
$M$ matters only at very small values: a group averages about 17 of a frame's 196 tokens, making $M{=}2$ insufficient and lowering \Ptwo accuracy.
From $M{=}4$ onward, accuracy changes little while memory grows; $M{=}16$ retains nearly all tokens in an average group, offering little gain over $M{=}8$.

\smallskip\noindent\textbf{Basis widths $D_U$ and $D_\Phi$.} \Cref{fig:s_sens}(c, d) varies the shared and group basis widths.
Very narrow bases ($D_U{=}12$, $D_\Phi{=}2$) lose details of the earlier state and lower \Ptwo, whereas wider shared bases change accuracy little.
A wider group basis keeps more of the variation within each group, which raises \Ptwo but lowers \Ctwo, and the two curves cross at our setting.
Keeping past frames in more detail may thus hinder reading the current state when it differs from the earlier one.
This trade-off appears only when both states are asked about at the same moment, as in \data, which makes it a useful testbed for future work on how much past detail a streaming memory should keep.
Wider bases also cost more storage, up to 21.8\% of raw KV against 5.8\% for ours.
All settings in the four sweeps keep \emph{Cross-state} above every LLaVA-OV baseline (Table~2 of the main paper), so our gain does not depend on a narrow choice of these settings.

\subsection{Grouping and Residual Correction}
\label{sec:s_design}
\begin{table}[t]
\centering
\footnotesize
\setlength{\tabcolsep}{2.7pt}
\caption{\textbf{Accuracy (\%) on \data by grouping feature and rule.} \#groups: fixed for Random and $k$-means; for BKM, the mean number of groups per past frame and layer. Mem.: stored records per past frame (MiB). Cons.: \emph{Consistency}. Shaded: ours.}

\label{tab:s_group}
\begin{tabular}{@{}l l c c c c c c@{}}
\toprule
\multirow{2}{*}{\hd{Feature}} & \multirow{2}{*}{\hd{Rule}} & \hd{\#} & \multirow{2}{*}{\hd{Mem.}} & \hd{$s_1$} & \hd{Cross-} & \hd{Cross-} & \multirow{2}{*}{\hd{Cons.}} \\[-1pt]
 & & \hd{groups} & & \hd{$@t_2$} & \hd{time} & \hd{state} & \\
\midrule
\multirow{3}{*}{Visual} & Random & 11.0 & 0.63 & 42.5 & 34.1 & 18.0 & 44.7 \\
 & $k$-means & 11.0 & 0.57 & 44.5 & 37.3 & 19.3 & 47.0 \\
 & BKM & 10.6 & 0.61 & 46.1 & 37.3 & 19.1 & 44.5 \\
\midrule
\multirow{5}{*}{KV} & Random & 11.0 & 0.63 & 41.4 & 33.9 & 16.8 & 42.3 \\
 & $k$-means & 4.0 & 0.14 & 43.6 & 35.5 & 19.3 & 46.2 \\
 & $k$-means & 11.0 & 0.56 & 47.3 & 38.4 & 20.9 & 47.3 \\
 & $k$-means & 16.0 & 0.97 & 48.9 & \textbf{40.2} & \textbf{22.7} & \textbf{49.7} \\
\rowcolor{oursrow}\cellcolor{white} & BKM & 11.3 & 0.60 & 47.3 & \underline{39.1} & \underline{22.0} & \underline{49.4} \\
\bottomrule
\end{tabular}
\end{table}

\smallskip\noindent\textbf{Grouping feature and rule.} \Cref{tab:s_group} compares grouping features and rules, with random groups as a reference.
Groups formed on the visual tokens are shared by all layers and fit the KV of each layer less well, which mainly lowers \Ptwo.
On the per-layer KV, $k$-means with a fixed $K$ trades accuracy for memory, since memory grows faster than $K$.
BKM adapts the number of groups to each frame and layer: with about the same mean number of groups as $K{=}11$, it is slightly better on the paired metrics and comes within about one point of $K{=}16$ at 0.6 times its memory.
\begin{table}[t]
\centering
\footnotesize
\setlength{\tabcolsep}{3.0pt}
\caption{\textbf{Effect of the extra residual direction $\boldsymbol{\phi}_g$ on \data.} Mem.: stored records per frame (MiB). Error: squared reconstruction error of the shared tokens.}
\label{tab:s_resid}
\begin{tabularx}{\linewidth}{@{}l *{5}{>{\centering\arraybackslash}X}@{}}
\toprule
 & & \multicolumn{2}{c}{\hd{Error$\downarrow$}} & \hd{Cross-} & \hd{Cross-} \\[-1pt]
\cmidrule(lr){3-4}
\hd{Method} & \hd{Mem.} & \hd{Key} & \hd{Value} & \hd{time} & \hd{state} \\
\midrule
w/o $\boldsymbol{\phi}_g$ & 0.51 & 0.664 & 0.799 & 38.9 & 21.6 \\
\rowcolor{oursrow}\textbf{Ours} & 0.60 & \textbf{0.530} & \textbf{0.718} & \textbf{39.1} & \textbf{22.0} \\
\bottomrule
\end{tabularx}
\end{table}

\smallskip\noindent\textbf{Extra residual direction.} A shared group reuses the bases of its parent, so part of its residual falls outside them; the extra direction $\boldsymbol{\phi}_g$ keeps the main direction of this part.
\Cref{tab:s_resid} shows that $\boldsymbol{\phi}_g$ lowers the reconstruction error of the shared keys and values at a small memory cost, while paired accuracy barely changes.
We therefore keep $\boldsymbol{\phi}_g$ as an inexpensive safeguard for groups that reuse a parent's statistics.

\begin{figure}[t]
\centering
\includegraphics[width=0.9\linewidth]{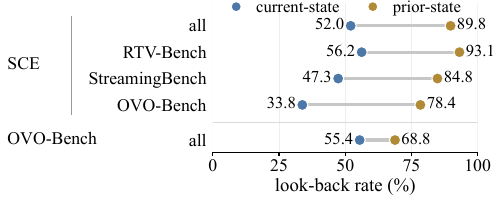}
\caption{\textbf{Look-back rate of the gate.} Percentage of current- and prior-state questions for which our gate looks back ($\tau{=}0.015$). On \data, \Cone and \Ctwo are current-state and \Ptwo is prior-state, shown for all cases and for each source. The bottom row is OVO-Bench itself, whose real-time (837) and backward-tracing (631) questions are current- and prior-state.}
\label{fig:s_lookback}
\end{figure}

\begin{figure*}[t]
\centering
\includegraphics[width=0.87\textwidth]{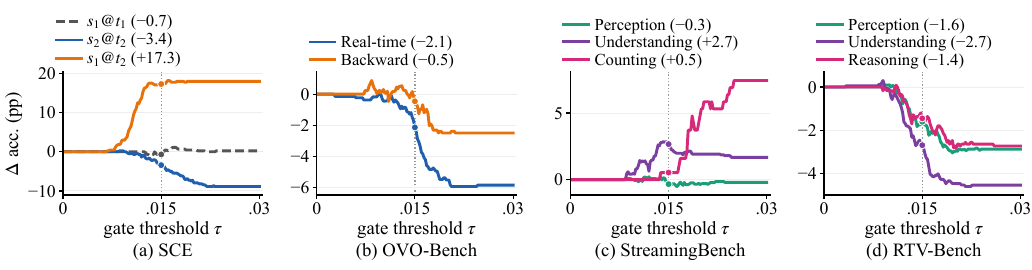}
\caption{\textbf{Gate threshold sensitivity.} Change in accuracy relative to $\tau{=}0$ (no look-back) up to $\tau{=}0.03$ (nearly always look back), by question type. Lines in (b)--(d): mean over tasks. Dots and values in parentheses: our setting ($\tau{=}0.015$). StreamingBench perception: object, attribute, and action perception and spatial understanding; understanding: causal reasoning, clips summarization, event understanding, prospective reasoning, and text-rich understanding. The RTV-Bench groups follow the benchmark’s official task categories.}
\label{fig:s_tau}
\end{figure*}

\subsection{Look-Back Gate}
\label{sec:s_gate}

\smallskip\noindent\textbf{Look-back rates.} \Cref{fig:s_lookback} shows that the gate looks back more often on prior-state than on current-state questions, both on \data and on OVO-Bench.
This holds for the backward-tracing questions of OVO-Bench, which lack the prefix ``Near the beginning of this clip'' of every \Ptwo question, so the gain of \ours does not stem from this prefix.
The gate still looks back on about half of the current-state questions, where retrieval can break a correct recent-only answer, so an oracle picking the better route exceeds it on every benchmark (Table~7 of the main paper), and more precise routing is a promising direction for future work.

\smallskip\noindent\textbf{Threshold.} \Cref{fig:s_tau} varies $\tau$ from 0, where no question looks back, to 0.03, where nearly every question does.
Our $\tau{=}0.015$ is not the best value on any benchmark.
On OVO-Bench and RTV-Bench, accuracy mostly drops as $\tau$ grows, in line with the finding that recent frames alone often suffice on existing benchmarks~\cite{simplestream2026,deltas2026}; StreamingBench is the exception, as its counting questions ask for counts accumulated over the stream.

\section{Per-Stage Latency}
\label{sec:s_mem}

\smallskip\noindent\textbf{Comparison setup.} We compare ours with the retrieval-based baselines, which also keep past frames to retrieve from, using their released code with the settings of \Cref{sec:s_impl}.
All methods run with LLaVA-OV on one RTX A6000 GPU, one method at a time, on 1,024 frames at 1\,fps.

\smallskip\noindent\textbf{Memory write.} In a stream, writing frames in chunks lowers the time per frame but adds a batching delay, since a frame cannot be retrieved until its chunk fills.
We therefore feed one frame per call at 1\,fps and divide the total time to write the same 1,024 frames by the number of frames.
ShallowStream's write time is overestimated because its released code concatenates each new frame onto all stored input embeddings at every call.
Ours writes a frame in 160\,ms, so each frame becomes retrievable well within the 1\,s between frames; given that it stores each frame at about $1/16$ of its raw KV, this write time remains competitive with the retrieval-based baselines.

\begin{figure}[t]
\centering
\begin{subfigure}{\linewidth}\centering
  \includegraphics[width=0.92\linewidth]{{{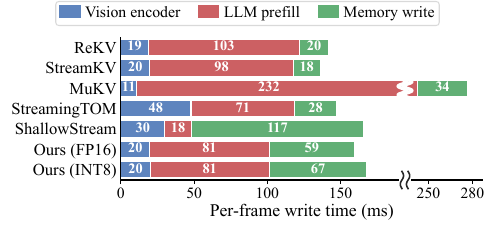}}}
  \caption{Per-frame write time}\label{fig:s_mem_write}
\end{subfigure}\\[1pt]
\begin{subfigure}{\linewidth}\centering
  \includegraphics[width=0.92\linewidth]{{{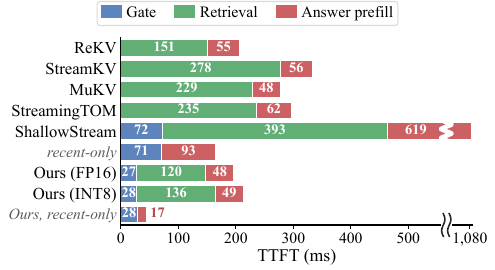}}}
  \caption{Time to first token}\label{fig:s_mem_ttft}
\end{subfigure}
\caption{\textbf{Per-stage latency.} (a) Per-frame write time and (b) median time to first token (TTFT) at 1,024 frames. Retrieval includes loading or decoding the retrieved KV, and memory write and retrieval include the remaining overhead. Gated methods show both the look-back and the recent-only routes.}
\label{fig:s_mem}
\end{figure}

\smallskip\noindent\textbf{Time to first token.} TTFT is measured on the memory that each method builds at its native chunk size.
The main paper overestimated the TTFT of MuKV and StreamKV because of our timing code; after the fix, they take 277 and 333\,ms.
ShallowStream's recent-only TTFT in the main paper is exaggerated by an unused scoring of all stored frames; without it, the route takes 165\,ms.
Our compact records stay on the GPU, so our retrieval loads nothing from the CPU and is the shortest among the retrieval-based methods; even with the gate, our look-back TTFT in FP16 is also the shortest (\Cref{fig:s_mem}), and our recent-only route answers in 45\,ms.

\section{Per-Task Results on Existing Benchmarks}
\label{sec:s_pertask}

\begin{table*}[p]
\centering
\caption{\textbf{Accuracy (\%) on OVO-Bench.} Real-time: OCR, ACR (action), ATR (attribute), STU (spatial), FPD (future prediction), OJR (object). Backward: EPM (episodic memory), ASI (action sequence), HLD (hallucination). Avg.: mean over the tasks of each category. Overall: mean of the two averages.}
\label{tab:ovo}
{\footnotesize\setlength{\tabcolsep}{3.6pt}\renewcommand{\arraystretch}{0.86}
\begin{tabularx}{\textwidth}{@{}l *{12}{>{\centering\arraybackslash}X}@{}}
\toprule
 & \multicolumn{7}{c}{\hd{Real-Time Visual Perception}} & \multicolumn{4}{c}{\hd{Backward Tracing}} & \\
\cmidrule(lr){2-8}\cmidrule(lr){9-12}
\hd{Method} & \hd{OCR} & \hd{ACR} & \hd{ATR} & \hd{STU} & \hd{FPD} & \hd{OJR} & \hd{Avg.} & \hd{EPM} & \hd{ASI} & \hd{HLD} & \hd{Avg.} & \hd{Overall} \\
\midrule
\rowcolor{bbrow}\textbf{LLaVA-OV-7B}~\cite{llavaonevision2025} & & & & & & & & & & & & \\
\rep{\plus SimpleStream~\cite{simplestream2026}} & \rep{\textbf{79.2}} & \rep{\underline{75.2}} & \rep{75.9} & \rep{\underline{57.3}} & \rep{\underline{74.3}} & \rep{\textbf{75.0}} & \rep{\underline{72.8}} & \rep{57.2} & \rep{56.8} & \rep{20.4} & \rep{44.8} & \rep{\underline{58.8}} \\
\famsep{13}
\rep{\plus HERMES~\cite{hermes2026}} & \rep{75.8} & \rep{64.2} & \rep{\textbf{80.2}} & \rep{49.4} & \rep{\textbf{75.2}} & \rep{\underline{74.5}} & \rep{69.9} & \rep{\underline{61.6}} & \rep{\textbf{63.5}} & \rep{25.8} & \rep{\underline{50.3}} & \rep{\textbf{60.1}} \\
\rep{\plus CausalMem~\cite{causalmem2026}} & \rep{75.2} & \rep{59.6} & \rep{73.3} & \rep{48.3} & \rep{\underline{74.3}} & \rep{66.8} & \rep{66.2} & \rep{59.3} & \rep{60.8} & \rep{12.9} & \rep{44.3} & \rep{55.3} \\
\rep{\plus CoRDS~\cite{cords2026}} & \rep{73.8} & \rep{65.1} & \rep{76.7} & \rep{46.6} & \rep{\textbf{75.2}} & \rep{65.2} & \rep{67.1} & \rep{\textbf{62.0}} & \rep{\underline{61.5}} & \rep{\underline{28.0}} & \rep{\textbf{50.5}} & \rep{\underline{58.8}} \\
\rep{\plus InfiniPot-V~\cite{infinipotv2025}} & \rep{72.5} & \rep{62.4} & \rep{78.4} & \rep{50.6} & \rep{\textbf{75.2}} & \rep{69.0} & \rep{68.0} & \rep{59.6} & \rep{60.8} & \rep{15.1} & \rep{45.2} & \rep{56.6} \\
\famsep{13}
\plus ReKV~\cite{rekv2025} & 67.1 & 54.1 & 72.4 & 47.8 & 72.3 & 64.7 & 63.1 & 59.6 & 59.5 & 19.4 & 46.1 & 54.6 \\
\plus StreamKV~\cite{streamkv2026} & 69.1 & 59.6 & 75.0 & 45.5 & 72.3 & 64.7 & 64.4 & 59.6 & 54.1 & 16.1 & 43.3 & 53.8 \\
\plus MuKV~\cite{mukv2026} & 51.0 & 47.7 & 62.1 & 37.1 & 67.3 & 44.0 & 51.5 & 50.5 & 53.4 & 23.7 & 42.5 & 47.0 \\
\plus StreamingTOM~\cite{streamingtom2026} & 61.7 & 52.3 & 74.1 & 46.6 & 69.3 & 57.6 & 60.3 & 55.2 & 57.4 & 14.0 & 42.2 & 51.2 \\
\plus ShallowStream~\cite{shallowstream2026} & \underline{76.5} & \textbf{78.9} & 78.4 & \textbf{59.0} & 72.3 & 73.9 & \textbf{73.2} & 50.8 & 53.4 & \textbf{28.5} & 44.2 & 58.7 \\
\rowcolor{oursrow}\plus \textbf{Ours} & 74.5 & 73.4 & \underline{79.3} & 54.5 & 73.3 & 73.4 & 71.4 & 57.9 & 60.1 & \textbf{28.5} & 48.8 & \textbf{60.1} \\
\midrule
\rowcolor{bbrow}\textbf{Qwen2.5-VL-7B}~\cite{qwen25vl2025} & & & & & & & & & & & & \\
\rep{\plus SimpleStream~\cite{simplestream2026}} & \rep{\textbf{81.9}} & \rep{\textbf{69.7}} & \rep{\underline{73.3}} & \rep{\underline{51.1}} & \rep{\textbf{70.3}} & \rep{\textbf{65.2}} & \rep{\textbf{68.6}} & \rep{44.4} & \rep{\underline{60.1}} & \rep{\underline{38.2}} & \rep{\underline{47.6}} & \rep{\underline{58.1}} \\
\famsep{13}
\rep{\plus HERMES~\cite{hermes2026}} & \underline{81.2} & 60.6 & 68.1 & \underline{51.1} & 66.3 & 60.9 & 64.7 & \textbf{53.2} & \textbf{61.5} & 24.2 & 46.3 & 55.5 \\
\famsep{13}
\plus ReKV~\cite{rekv2025} & 79.2 & 56.9 & \textbf{75.0} & \textbf{51.7} & \textbf{70.3} & 61.4 & 65.7 & \underline{51.5} & 59.5 & 22.6 & 44.5 & 55.1 \\
\rowcolor{oursrow}\plus \textbf{Ours} & \underline{81.2} & \underline{63.3} & \textbf{75.0} & \textbf{51.7} & \underline{69.3} & \underline{63.0} & \underline{67.3} & 49.8 & 58.1 & \textbf{42.5} & \textbf{50.1} & \textbf{58.7} \\
\bottomrule
\end{tabularx}
}

\vspace{2pt}
\caption{\textbf{Accuracy (\%) on the StreamingBench real-time split.} OP: object perception, CR: causal reasoning, CS: clips summarization, ATP: attribute perception, EU: event understanding, TR: text-rich understanding, PR: prospective reasoning, SU: spatial understanding, ACP: action perception, CT: counting. Avg.: mean over the ten tasks.}
\label{tab:sb}
{\footnotesize\setlength{\tabcolsep}{3.6pt}\renewcommand{\arraystretch}{0.86}
\begin{tabularx}{\textwidth}{@{}l *{11}{>{\centering\arraybackslash}X}@{}}
\toprule
\hd{Method} & \hd{OP} & \hd{CR} & \hd{CS} & \hd{ATP} & \hd{EU} & \hd{TR} & \hd{PR} & \hd{SU} & \hd{ACP} & \hd{CT} & \hd{Avg.} \\
\midrule
\rowcolor{bbrow}\textbf{LLaVA-OV-7B}~\cite{llavaonevision2025} & & & & & & & & & & & \\
\rep{\plus SimpleStream~\cite{simplestream2026}} & \rep{81.6} & \rep{71.9} & \rep{\underline{87.7}} & \rep{\textbf{84.0}} & \rep{72.3} & \rep{73.8} & \rep{73.1} & \rep{\underline{69.1}} & \rep{\textbf{75.6}} & \rep{31.9} & \rep{72.1} \\
\famsep{12}
\rep{\plus HERMES~\cite{hermes2026}} & \rep{\underline{81.8}} & \rep{82.0} & \rep{\textbf{89.9}} & \rep{\underline{83.0}} & \rep{\textbf{74.8}} & \rep{\underline{76.0}} & \rep{\underline{74.1}} & \rep{67.5} & \rep{70.7} & \rep{42.6} & \rep{\textbf{74.2}} \\
\rep{\plus CausalMem~\cite{causalmem2026}} & \rep{79.9} & \rep{\underline{82.8}} & \rep{87.4} & \rep{80.4} & \rep{69.8} & \rep{\textbf{76.6}} & \rep{71.3} & \rep{64.2} & \rep{67.9} & \rep{32.4} & \rep{71.3} \\
\rep{\plus CoRDS~\cite{cords2026}} & \rep{81.6} & \rep{82.0} & \rep{87.1} & \rep{82.1} & \rep{\underline{73.6}} & \rep{73.8} & \rep{71.3} & \rep{\textbf{69.5}} & \rep{70.7} & \rep{42.0} & \rep{\underline{73.4}} \\
\rep{\plus InfiniPot-V~\cite{infinipotv2025}} & \rep{\textbf{83.2}} & \rep{\underline{82.8}} & \rep{85.2} & \rep{81.4} & \rep{72.3} & \rep{75.7} & \rep{70.4} & \rep{66.7} & \rep{69.3} & \rep{31.4} & \rep{71.8} \\
\famsep{12}
\plus ReKV~\cite{rekv2025} & 77.2 & 79.7 & 78.2 & 77.6 & 69.8 & 69.2 & 69.4 & 63.0 & 64.5 & 43.6 & 69.2 \\
\plus StreamKV~\cite{streamkv2026} & 79.1 & 80.5 & 82.3 & 79.8 & 68.6 & 68.8 & 73.1 & 61.0 & 65.6 & \underline{44.7} & 70.4 \\
\plus MuKV~\cite{mukv2026} & 67.5 & 73.4 & 71.0 & 67.0 & 61.6 & 57.3 & 58.3 & 56.1 & 58.8 & 33.5 & 60.5 \\
\plus StreamingTOM~\cite{streamingtom2026} & 76.4 & \textbf{83.6} & 79.2 & 73.7 & 66.0 & 67.0 & 68.5 & 59.8 & 62.8 & \textbf{45.7} & 68.3 \\
\plus ShallowStream~\cite{shallowstream2026} & 79.7 & 68.8 & 85.5 & 80.4 & 71.7 & 72.0 & 70.4 & 68.7 & \underline{73.9} & 35.1 & 70.6 \\
\rowcolor{oursrow}\plus \textbf{Ours} & 81.3 & 80.5 & 86.8 & 79.8 & \underline{73.6} & 75.1 & \textbf{76.9} & 67.5 & 71.3 & 39.4 & 73.2 \\
\midrule
\rowcolor{bbrow}\textbf{Qwen2.5-VL-7B}~\cite{qwen25vl2025} & & & & & & & & & & & \\
\rep{\plus SimpleStream~\cite{simplestream2026}} & \rep{74.3} & \rep{68.0} & \rep{\textbf{82.6}} & \rep{81.4} & \rep{68.6} & \rep{\underline{78.8}} & \rep{76.9} & \rep{\textbf{71.5}} & \rep{\textbf{71.9}} & \rep{38.8} & \rep{71.3} \\
\famsep{12}
\rep{\plus HERMES~\cite{hermes2026}} & \rep{\underline{78.0}} & \rep{\textbf{78.9}} & \rep{81.1} & \rep{\underline{83.7}} & \rep{\underline{72.3}} & \rep{78.2} & \rep{77.8} & \rep{\underline{66.7}} & \rep{67.3} & \rep{\textbf{58.0}} & \rep{\textbf{74.2}} \\
\famsep{12}
\plus ReKV~\cite{rekv2025} & 76.4 & \underline{78.1} & 80.4 & 82.1 & 71.7 & 77.9 & \underline{81.5} & 62.2 & 67.9 & \underline{52.1} & 73.0 \\
\rowcolor{oursrow}\plus \textbf{Ours} & \textbf{78.3} & 75.0 & \underline{81.4} & \textbf{84.0} & \textbf{73.6} & \textbf{81.0} & \textbf{82.4} & 65.4 & \underline{69.3} & 43.1 & \underline{73.4} \\
\bottomrule
\end{tabularx}
}

\vspace{2pt}
\caption{\textbf{Accuracy (\%) on RTV-Bench.} Perception: TP (temporal), VP (visual), SP (scene); Understanding: GU (global), PU (phenomenological), IA (intent analysis); Reasoning: FP (future prediction), SR (spatiotemporal reasoning).}
\label{tab:rtv}
{\footnotesize\setlength{\tabcolsep}{3.6pt}\renewcommand{\arraystretch}{0.86}
\begin{tabularx}{\textwidth}{@{}l *{9}{>{\centering\arraybackslash}X}@{}}
\toprule
 & \multicolumn{3}{c}{\hd{Perception}} & \multicolumn{3}{c}{\hd{Understanding}} & \multicolumn{2}{c}{\hd{Reasoning}} & \\
\cmidrule(lr){2-4}\cmidrule(lr){5-7}\cmidrule(lr){8-9}
\hd{Method} & \hd{TP} & \hd{VP} & \hd{SP} & \hd{GU} & \hd{PU} & \hd{IA} & \hd{FP} & \hd{SR} & \hd{Overall} \\
\midrule
\rowcolor{bbrow}\textbf{LLaVA-OV-7B}~\cite{llavaonevision2025} & & & & & & & & & \\
\rep{\plus SimpleStream~\cite{simplestream2026}} & \rep{\underline{58.9}} & \rep{\textbf{53.9}} & \rep{55.5} & \rep{47.7} & \rep{\underline{62.4}} & \rep{\underline{60.5}} & \rep{\underline{60.1}} & \rep{\underline{47.1}} & \rep{\underline{55.8}} \\
\famsep{10}
\rep{\plus HERMES~\cite{hermes2026}} & \rep{53.3} & \rep{51.7} & \rep{53.1} & \rep{47.3} & \rep{60.4} & \rep{57.8} & \rep{57.3} & \rep{42.7} & \rep{52.9} \\
\rep{\plus CausalMem~\cite{causalmem2026}} & \rep{53.2} & \rep{50.1} & \rep{50.8} & \rep{45.4} & \rep{57.6} & \rep{52.5} & \rep{52.7} & \rep{43.9} & \rep{50.8} \\
\rep{\plus CoRDS~\cite{cords2026}} & \rep{50.9} & \rep{50.7} & \rep{47.9} & \rep{40.9} & \rep{52.8} & \rep{45.8} & \rep{47.3} & \rep{43.5} & \rep{47.7} \\
\rep{\plus InfiniPot-V~\cite{infinipotv2025}} & \rep{54.7} & \rep{49.5} & \rep{51.2} & \rep{45.8} & \rep{57.4} & \rep{52.2} & \rep{56.8} & \rep{43.4} & \rep{51.4} \\
\famsep{10}
\plus ReKV~\cite{rekv2025} & 49.1 & 48.6 & 48.0 & 44.5 & 52.7 & 48.7 & 51.2 & 40.1 & 47.9 \\
\plus StreamKV~\cite{streamkv2026} & 49.2 & 48.3 & 48.4 & 42.6 & 54.9 & 49.8 & 52.6 & 41.0 & 48.4 \\
\plus MuKV~\cite{mukv2026} & 43.8 & 46.5 & 41.9 & 38.8 & 44.8 & 42.8 & 46.8 & 35.2 & 42.8 \\
\plus StreamingTOM~\cite{streamingtom2026} & 49.6 & 46.7 & 47.1 & 42.4 & 52.8 & 47.5 & 49.8 & 37.2 & 46.7 \\
\plus ShallowStream~\cite{shallowstream2026} & \textbf{59.8} & \underline{52.4} & \textbf{57.5} & \textbf{50.0} & \textbf{63.7} & \textbf{61.4} & \textbf{61.9} & \textbf{48.5} & \textbf{56.8} \\
\rowcolor{oursrow}\plus \textbf{Ours} & 53.0 & 52.0 & \underline{55.7} & \underline{48.1} & 59.4 & 56.7 & 57.7 & 45.4 & 53.5 \\
\midrule
\rowcolor{bbrow}\textbf{Qwen2.5-VL-7B}~\cite{qwen25vl2025} & & & & & & & & & \\
\rep{\plus SimpleStream~\cite{simplestream2026}} & \rep{\textbf{60.1}} & \rep{54.5} & \rep{\textbf{54.6}} & \rep{48.5} & \rep{\textbf{64.4}} & \rep{\textbf{62.3}} & \rep{\textbf{57.5}} & \rep{45.4} & \rep{\textbf{56.0}} \\
\famsep{10}
\rep{\plus HERMES~\cite{hermes2026}} & \rep{54.2} & \rep{\textbf{57.1}} & \rep{52.9} & \rep{\textbf{50.2}} & \rep{58.9} & \rep{54.3} & \rep{\underline{55.1}} & \rep{\underline{46.6}} & \rep{53.8} \\
\famsep{10}
\plus ReKV~\cite{rekv2025} & 52.5 & 55.6 & 49.5 & 46.0 & 56.5 & 54.0 & 49.0 & 45.1 & 51.2 \\
\rowcolor{oursrow}\plus \textbf{Ours} & \underline{57.9} & \underline{56.0} & \underline{53.3} & \underline{48.7} & \underline{61.5} & \underline{56.7} & 52.6 & \textbf{47.8} & \underline{54.4} \\
\bottomrule
\end{tabularx}
}

\end{table*}

\Cref{tab:ovo,tab:sb,tab:rtv} give the per-task results on OVO-Bench~\cite{ovobench2025}, StreamingBench~\cite{streamingbench2026}, and RTV-Bench~\cite{rtvbench2025}, with all methods rerun in our pipeline.
Most of their questions concern the current state, so SimpleStream and ShallowStream, which answer mostly from the recent frames, lead on OVO-Bench real-time perception and on RTV-Bench, while retrieval-based methods that always look back score lower except on StreamingBench counting, which asks for accumulated counts.
Ours remains competitive, with the highest overall score on OVO-Bench and the highest score on RTV-Bench among the methods that keep and use a memory of the past.
However, these benchmarks score each question on its own, so they do not test whether a streaming memory keeps both the prior and the current state.
This highlights the need for a benchmark such as \data, which pairs current- and prior-state questions on each state change.

\end{document}